\documentclass[pdflatex,sn-apa]{sn-jnl}
\jyear{2022}
\theoremstyle{thmstyleone}%
\theoremstyle{thmstyletwo}%

\theoremstyle{thmstylethree}%

\begin{document}

\title[Source data selection for out-of-domain generalization]{Source data selection for out-of-domain generalization}

\author[1]{\fnm{Xinran} \sur{Miao}}\email{xmiao27@wisc.edu}

\author*[1]{\fnm{Kris} \sur{Sankaran}}\email{ksankaran@wisc.edu}

\affil[1]{\orgdiv{Department of Statistics}, \orgname{University of Wisconsin - Madison}, \orgaddress{\street{1300 University Ave}, \city{Madison}, \postcode{53706}, \state{WI}}}

\abstract{Models that perform out-of-domain generalization borrow knowledge from heterogeneous source data and apply it to a related but distinct target task. Transfer learning has proven effective for accomplishing this generalization in many applications. 
However, poor selection of a source dataset can lead to poor performance on the target, a phenomenon called negative transfer. In order to take full advantage of available source data, this work studies source data selection with respect to a target task. We propose two source selection methods that are based on the multi-bandit theory and random search, respectively. 
We conduct a thorough empirical evaluation on both simulated and real data. Our proposals can be also viewed as diagnostics for the existence of a reweighted source subsamples that perform better than the random selection of available samples.}

\keywords{Out-of-domain generalization, Domain shift, Source selection, Transfer Learning, Multi-armed bandit}

\maketitle

\newpage
\section{Introduction}\label{sec:introduction}
In many modern prediction modeling applications, the operating conditions of an algorithm may not match the setting in which it was trained. For example, this may result from a change in spatial or temporal context. In either case, ephemeral predictors — those present in the training data, but not in the target setting — can result in performance drops \citep{tsymbal2004problem,rosenstein2005transfer, zhang2020overcoming}.
This problem is especially acute when labeled data scarce in the target regime. As a way of borrowing extra knowledge from out-of-distribution source data, transfer learning shows great performance in practice, but its mysterious mechanism makes it difficult to choose a suitable dataset to pretrain a model on, especially under a computational budget \citep{wang2019characterizing}. 
This is referred to as the out-of-domain generalization problem, and it has attracted substantial interest from the community over the last two decades \citep{moreno2012unifying, li2018learning, yue2019domain, recht2019imagenet}. Important advances include the use of transfer learning methods to improve performance on small target datasets \citep{long2013transfer,ghifary2014domain,guo2018deep} and the introduction of regularization strategies that encourage models to learn structure that generalizes across contexts \citep{kirkpatrick2017overcoming,chen2019transferability, chen2019catastrophic}.

However, existing approaches are difficult to interpret. When a model performs much worse outside of the context that it was trained, it can be difficult to isolate characteristics of the training data or resulting model that are responsible for the deterioration, which is called negative transfer \citep{rosenstein2005transfer, wang2019characterizing}. Conversely, performance gains may be obtained when pretraining models on different source datasets, but the mechanisms are often unclear.
For purely automated systems, prediction performance is still the key property. However, in an increasing number of scientific or social applications, models must also be used by domain experts to better understand their systems of interest, and in this context, attribution is critical.

Given target and source tasks, one can always supplement data from the target task with data from the source. The question is: Does the choice make a difference, and if so, can we understand and take advantage of it?
To address this, we empirically explore several strategies for training models across non-identically distributed subsets of a source dataset and measuring how model performance varies across reweighted sources. We consider both linear regression models trained on tabular and geographic data and deep learning models trained on imagery of pathology slides. 
Specifically, we investigate ensemble and bandit-based approaches for highlighting subsets of the source data that strongly affect performance on the target. Differential performance across ensemble members or a strong preference for certain bandit arms are used as indicators of distributional heterogeneity. 
Further, we propose accompanying visualization strategies that clarify the relationship between source and target data within these algorithms, identifying the optimal source compositions of the ensemble method and the trajectories of the bandit selection method. 
Finally, we study in detail the factors that influence these source selection methods, including the initial data representation, the criteria for splitting the sources, and the relative abundance of data similar to the target.

In section \ref{sec:background} we highlight commonly used approaches to out-of-domain generalization and then review concepts that directly inform our approach. In section \ref{sec:methods}, we detail our source selection strategies. In section \ref{sec:simulation}, we apply these methods to a simulation reflecting the temporal generalization problem. Section \ref{sec:data_analysis} presents two case studies, one of home prices across California and another on tumor identification across hospitals with different patient populations and imaging equipment.
We note that all data and code described below are available: \url{https://github.com/XinranMiao/source_selection}.

\section{Background}
\label{sec:background}

\subsection{Transfer and negative transfer}
Consider a prediction task with target data of interest $\mathcal{D}^t = \{x_{i}^t, y_{i}^t\}_{i=1}^{n^t}$, where $x_{i}^t$'s and $y_{i}^t$'s are the observations of variables and responses and $n^t$ is the target data size. 
When $n^t$ is small and a related larger supplementary dataset is available, a common rule-of-thumb is to pretrain the model on this larger dataset before training it on the target. This practice is called is transfer learning \citep{torrey2010transfer}. 
Denote this related source dataset by $\mathcal{D}^s = \{x_i^s,y_i^s\}_{i=1}^{n^s}$ where $n^s \gg n^t$. 
We will often have metadata about each observation in the source. This can be information aside from the predictors or certain data representations that we are unavailable in the target. Let $p^t$ and $p^s$ be the unknown distributions of target and source, and $\theta \in \Theta$ be the parameter of the prediction model. Then transfer learning learns a model $f_\theta $ based on $\mathcal{D}^s$, 
which can be fine-tuned on $D^t$ and ideally yields $f_\theta(x_{\ast}^t) \approx y_{\ast}^t$ for new samples $\left(x_{\ast}^t, y_\ast^t\right) \sim p^t$.

Despite the popularity of transfer learning, negative transfer appears -- training models on source together with target data can be worse than training on target data alone \citep{rosenstein2005transfer, wang2019characterizing, zhang2020overcoming}. This can be caused by distributional gaps \citep{koh2020wilds}, where $p^t$ and $p^s$ are very dissimilar. For example, satellite images from Africa and the Arctic Pole can be extremely different for landcover segmentation due to dissimilar feature spaces and prediction rules. Note that both the covariate distribution and the conditional distribution of response given covariates may change. \citep{zhang2020overcoming} summarizes methods to overcome negative transfer, including selecting or reweighting source data \citep{yao2010boosting,li2019multisource}, improving target data labeling \citep{gholami2020unsupervised}, addressing domain divergence \citep{shen2018wasserstein}, and integrated approaches \citep{kuzborskij2013stability}. 

Instead of working on fixed source and target sets, there also have been attempts to investigate the transferability of source datasets across an entire application area.
In these studies, the goal is to identify a small number of source datasets that can be used for transfer learning across a variety of problems. 
For example,  \citep{neumann2020training} explores qualitatively important factors in the remote sensing context, identifying source datasets whose learned representations lead to high performance on various downstream tasks. 

\subsection{Subset selection}

With the purpose of saving computation and avoiding negative transfer, researchers have developed subset selection strategies to weight the source dataset effectively with respect to the target task. Suppose we partition the source into $K$ subsets $\mathcal{D}^{s}_1,\mathcal{D}^{s}_2,\cdots,\mathcal{D}^{s}_K$, with empirical distributions $\hat{p}^s_1,\cdots,\hat{p}^s_K$. Let $\*w = [w_1,\cdots,w_K]$ be the weights of $K$ subsets, then with a specified loss function $L: \mathcal{D}^t\times \Theta \to \mathcal{R}^+$, we wish to minimize the generalization risk
\begin{align}
\min_{\*w \in \Delta^K, }\mathbb{E}_{p^t}[L(\mathcal{D}^t, \hat{\theta}\left(\*w\right))],
\label{formula:optimization}
\end{align}
where the model parameters are fit using
\begin{align}
\label{formula:estimator}
    \hat{\theta}\left(\*w\right) = \arg\min_{\theta \in \Theta} \mathbb{E}_{\hat{p}(\*w)}\left[L((x^s,y^s),\theta)\right]
\end{align} and $\hat{p}(\*w) = \sum_{k=1}^Kw_k \hat{p}^s_k$ is the weighted empirical subsampling distribution. The second minimization is the empirical risk estimator on the $\*w$-reweighted collection of source datsets. 
The first minimization searches across reweightings to minimize the loss on a dataset $\mathcal{D}^t$ drawn from the target $p^t$.

Sequential subset selection methods have been proposed under this framework. \cite{bouneffouf2014contextual} treat training examples individually as partitions. Let $\delta_{x_i^s}$ be the point mass at $x_i^s$. Then the weighted empirical subsampling distribution is $\frac{1}{n^s}\sum_{i=1}^{n^S}\delta_{x_i^s}$, which is a special case of $\hat{p}\left(\*w\right)$ with $K = n^s$, $\hat{p}_i^s = \delta_{x_i^s}$ and $w_i = 1/n^s$. They model the effect of adding one example onto the current training set by influence functions and then add the example with largest effect to the training set at each iteration. 
\cite{gutierrez2017multi} formulates this problem in a more general source partitioning setting and optimize it using Beta-Bernoulli Thompson Sampling, which sequentially adds source samples from different partitions with the choices of partitions updated by target performance. In this case, the weights $\*w$ are updated sequentially as a heuristic to optimizing (\ref{formula:optimization}). They demonstrate efficiency in training data selection method on medical image data. Our approach \ref{subsec:ts} translates this method into the context where source clusters are learned using alternative representations.
Instead of combining samples from different partitions to train a final model, \citep{yao2010boosting} trains weak classifiers sequentially into a stronger one using Boosting. Each weak classifier is trained by a weighted combination of one source partition and the target, with choice of partition decided by performance and weights updated iteratively. Here the weights consist of two numbers indicating one source partition and the target, which is different from $\*w$ introduced before.
Despite progress on subset selection algorithms, their dynamics can be difficult to characterize and diagnostic evaluation is not readily available. 
We build on these studies to further investigate performance attribution onto source subsets for out-of-domain generalization problems.


\section{Methods}
\label{sec:methods}
\subsection{Ensemble Method}
\label{subsec:ensemble}
The optimizer $(\hat{\*w},\hat{\theta})$ of (\ref{formula:optimization}) cannot be obtained analytically. One approach to obtaining an approximate empirical solution as described in Algorithm \ref{alg:ensemble}, which we call the ensemble method. 
For a given source dataset with $K$ subsets, we randomly sample subsets of size $n_{\text{training}}$ repeatedly for $J$ times, each time re-weighting the $K$ subsets differently. 
Specifically, in repetition $j$, we sample weights $\*w^j$ from a Dirichlet distribution with parameter $\mathbf{1}_K$, and then sample data $d^j$ from $\mathcal{D}^s$ that are weighted accordingly. 
After constructing the reweighted data, we train a model and evaluate it on the target task, yielding loss $l_j$. After $J$ repetitions, we find the $j$ with minimum $l_j$. The empirical optimizer would be the $\*w$ and $\hat{\theta}$ found in the $j^{th}$ interation, i.e., $(\hat{\*w},\hat{\theta} ) = (\*w^{\arg\min_{j}l_j},\hat{\theta}^{\arg\min_{j}l_j})$.
\begin{algorithm}
\caption{Sample weighting: Ensemble method}\label{alg:ensemble}
\begin{algorithmic}
\Require Number of source partitions $K$, number of repetitions $J$, and number of examples in each subsample $n_{\text{training}}$
\For{$j$ in $1:J$}
\State Sample weights $\*w^j = [w^j_1,w^j_2,\cdots,w^j_K] \sim \text{Dirichlet}(\mathbf{1}_K)$ for $K$ subsets.
\State For each $k$, sample $n_{\text{training}}\times w_{k}$ points from $\mathcal{D}^{s}_{k}$ to form $d_{k}^{j}$.
\State Fit the model $\hat{\theta}\left(w^{j}\right) = \min_{\theta \in \Theta} \sum_{k = 1}^{K}\sum_{\left(x_i, y_i\right) \in d_{k}^{j}} L\left(\left(x_i^t, y_i^t\right), \theta\right)$
\State Evaluate the target loss $l_j =  \sum_{i=1}^{n^t}L\left(\left(x_i^t, y_i^t\right), \hat{\theta}\left(w^{j}\right)\right)$
\EndFor
\State The optimizer is $(\hat{\*w},\hat{\theta} )= \left(\*w^{j^\ast}, \hat{\theta}\left(\*w^{j^\ast}\right)\right)$ where $j^{\ast} = \arg \min_{j} l_{j}$
\end{algorithmic}
\end{algorithm}

\subsection{Thompson Sampling with the Beta-Bernoulli Bandit}
\label{subsec:ts}
The multi-armed bandit problem deals with the situation in which an operator iteratively chooses one of a set of unknown distributions and observes a reward, with the goal of maximizing the cumulative gain \citep{lattimore2020bandit}. An example comes from a gambler at a row of slot machines, who has to decide which machine to play at each iteration. This setting is analogous to the sequential subset selection problem if we view different source subsets as arms in the bandit context.

Consider models trained on each source subset, and denote current parameter estimates of the reward distributions for each subset by $\lambda_1,\cdots,\lambda_K$. Starting from a randomly initialized model, at each iteration $h$, we add a batch of samples from one subset $\mathcal{D}_{k_h}$ and observe the reward $r_h$, with the purpose of maximizing the cumulative expected reward after $H$ rounds:
\begin{align}
\label{formula:ts}
    R_H = H\max_h\mathbb{E}[r_h] - \sum\limits_{h=1}^H \mathbb{E}[r_h].
\end{align}
Let reward $r_h$ be the indicator of whether adding samples from subset $k_h$ increases the target performance: 
\begin{align*}
 r_h = \begin{cases}
 0,& \text{if adding data from source } k_h \text{increases loss}\\
 1,& \text{otherwise},
 \end{cases}
\end{align*}

where $k_h \in \{1,2,\cdots,K\}$ indexes the source subset sampled at iteration $h$. We model the probability of this 0-1 reward at the $k^{th}$ source subset using $r_{h} \sim \mathrm{Ber}\left(\lambda_{k_{h}}\right)$. Assume priors $\lambda_k \sim \mathrm{Beta} \left(\alpha_{k}, \beta_{k}\right)$. Then the posterior for $\lambda_{k_h}$ at iteration $h$ can be updated to a Beta distribution with parameters $\alpha_{k_{h-1}}+\mathbb{I}[r_h = 1]$ and $\beta_{k_{h-1}}+\mathbb{I}[r_h = 0]$. At each iteration $h$, the choice of source subset $D_{k_h}$ is obtained by sampling $\hat{\lambda}_k$'s from current posteriors and choosing the arm with the highest probability of reward, i.e., $k_{h} = \arg\max_{k=1}^K \hat{\lambda}_k$, where $\hat{\lambda}_k \sim \mathrm{Beta}(\alpha_{k},\beta_{k}),\ k = 1,2,\cdots,K$. Algorithm \ref{alg:ts} provides the pseudocode of this Thompson Sampling with Beta-Bernoulli Bandit method. This approach addresses the trade-off between exploiting what is known and exploring new potentially useful source data.

\begin{algorithm}
\caption{Source selection: Thompson Sampling with Beta-Bernoulli Bandit}\label{alg:ts}
\begin{algorithmic}
\Require initial hyperparameters $\alpha_k=\alpha$, $\beta_k = \beta$, source subsets $\left\{\mathcal{D}_{k}\right\}_{k = 1}^{K}$, convergence error $\varepsilon>0$ 
\State Randomly initialize model parameter $\hat{\theta}$
\State Evaluate the model on the target set and obtain accuracy $a_0$
\While{$\lvert a_{h} - a_{h-1}\rvert > \varepsilon$}
\State $k_h = \arg\max_{k=1}^K\hat{\lambda}_k$, where $\hat{\lambda}_k\sim \text{Beta}(\alpha_k,\beta_k)$
\State Randomly select a batch of samples from $\mathcal{D}_{k_h}$, add them to the indices of training examples $I_{\text{training}}$
\State Update $\hat{\theta} = \arg\min_{\theta\in\Theta} \sum_{\left(x_i, y_i\right) \in I_{\text{training}}} L((x_i^t,y_i^t), \theta)$
\State Compute the prediction accuracy $a_{h}$ of the updated model on target
\If{$a_h>a_{h-1}$}
\State $\alpha_{k_h} = \alpha_{k_h}+1$
\Else
\State $\beta_{k_h} = \beta_{k_h}+1$
\EndIf
\EndWhile{}
\end{algorithmic}
\end{algorithm}

\subsection{Summary statistic}
\label{subsec:measure}

The necessity of subset selection depends on whether reweighting source subsets has any effect. We summarise the extent to which can this dataset be reduced to an efficient subset by computing the difference of optimal weights away from the uniform case
\begin{align}
   D =  \frac{1}{K}\sum_{k=1}^K \lvert\hat{w}_k - \frac{1}{K}\rvert.
\end{align}

\subsection{Protocols}
\label{subsec:protocols}
Generalization failures are typically caused by ephemeral predictors or distributional shifts \citep{wang2019characterizing}. We mimic these situations by splitting the dataset into source and target sets either using metadata information (section \ref{sec:simulation} and \ref{subsec:california}) or data representations (\ref{subsec:camelyon17}). After defining the source and target, we further split the source into $K$ subsets by either pre-defining some latent variable (section \ref{sec:simulation}) or clustering features (section \ref{subsec:california}) and/or their representations (section \ref{subsec:camelyon17}). Of the two splitting protocols, the former fits scenarios where data are collected at different times and locations with inherent differences, such as satellite images across continents or stock prices over a long period. Our goal is to determine whether, and possibly why, certain subsets make better (or worse) generalizations. The latter can be adopted in a more general situation without clear information on how distribution varies. In that case, the goal is more exploratory. 

Details of splitting the data are introduced in section \ref{sec:experiment} and vary among tasks and datasets. Given the target $\mathcal{D}^t$ and $K$ source subsets $\mathcal{D}^{s}_1,\mathcal{D}^{s}_2,\cdots,\mathcal{D}^{s}_K$, we explore the generalization problem using both the ensemble (section \ref{subsec:ensemble}) and the bandit-selection approaches (section \ref{subsec:ts}). We analyze experimental results both quantitatively (section \ref{subsec:measure}) and qualitatively.
\begin{figure}
    \centering
    \includegraphics[width=0.8\textwidth]{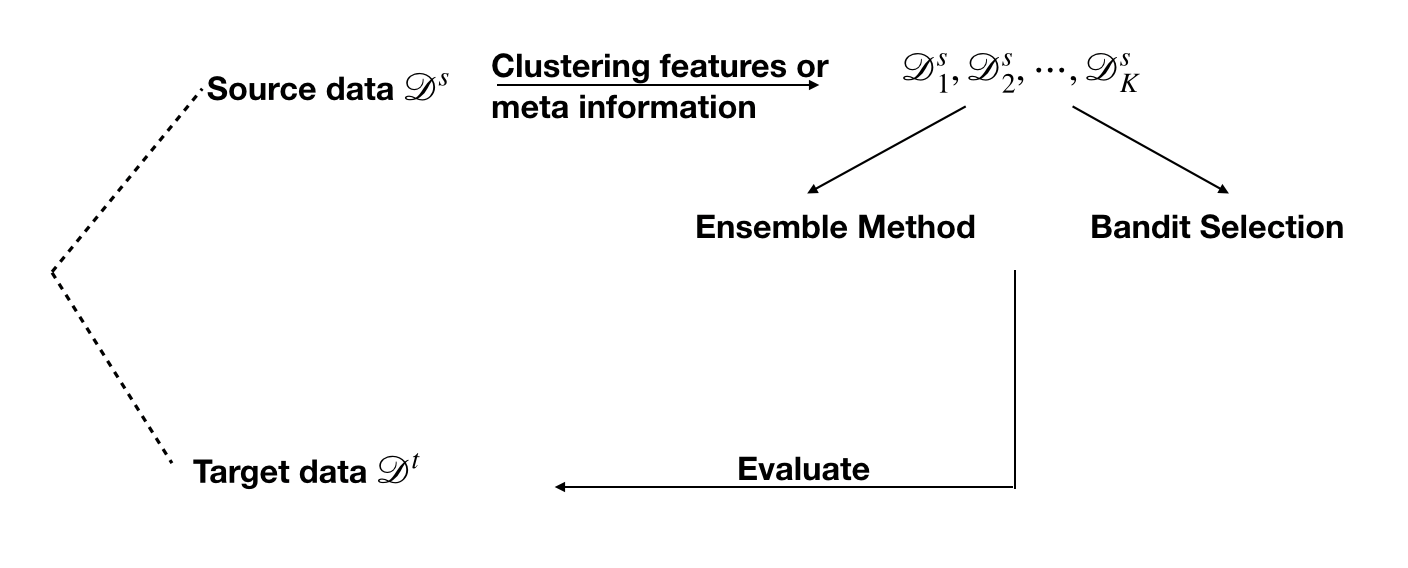}
    \caption{The workflow of experiment protocols.}
    \label{fig:chart_protocol}
\end{figure}


\section{Simulations}
\label{sec:simulation}

In temporal problems, predictors can be ephemeral, meaning that there is hidden context that induces changes in the underlying distribution  over time \citep{tsymbal2004problem}. It is therefore important to keep the model updated. Properly weighting data according to time can help avoid problems brought by epheremerality while still using of as much historical information as possible.

As an illustration, we simulate a dataset of size $1000$ where both the response $y\in \mathcal{R} $ and predictor $x\in\mathcal{R} ^4$ are collected over time. We consider time as the meta information $z$. Our target task is to make predictions on the most recent data (observations with large $z$ values) given the past. The relationship between $y$ and $x$ depends on $z\in (0,10]$:
\begin{align}
\label{formula:time}
  y =b_z^{T}x  + \varepsilon,
\end{align}

 where $b_z$ is a time-varying coefficient defined by formula (\ref{formula:B_z}) and $\varepsilon\sim \mathcal{N}\left(0,0.1\right)$. 

We take $x \sim \mathcal{N}\left(0, I_{p}\right)$. The coordinates of $b_{z}$ are generated by
 \begin{align}
\label{formula:B_z}
    b_{z,j} = \begin{cases}
    \beta_j  \alpha_j z + \varepsilon_j ,&0<z\leq 3 \\
   \beta_j (\alpha_j+1) z^2 + \varepsilon_j , &3<z \leq 5 \\
    \beta_j(\alpha_j -1)z+ \varepsilon_j , & 5 < z \leq 10,
    \end{cases}
\end{align}
where $\beta = [\beta_1,\beta_2,\beta_3,\beta_4]^T=[.9, .2, -.3, .3]^T$, $\varepsilon_j$'s are gaussian with mean zero and standard deviation $.01$, $.1$, $.04$, $.1$, and $\alpha_j$ follows a uniform distribution in $[-1,1]$. The interpretation that the coefficients in equation (\ref{formula:time}) are evolving linearly for the first three timepoints, quadratically for the next two, and linearly again for the remaining timepoints (but with a different slope).

We wish to investigate whether there exists a training subset weighted by time-based clusters that predicts the target better than a balanced subset. The training model is linear regression. We split source into $K=3$ subsets according to time $z$, and apply the bandit-selection and ensemble methods. The experiment follows from the protocol in section \ref{subsec:protocols} and values of parameters are listed in Table \ref{tab:simulation_para}.

\begin{table}[htp]
\caption{Model parameters of Simulation}
\label{tab:simulation_para}
\begin{tabular}{@{}llll@{}}
\hline
Sampling Procedure & Notation & Description & Value \\
\hline
{Both methods} & $K$ & Number of clusters in source & $3$ \\[6pt]
{Ensemble method} & $n_{\mathrm{training}}$ & Subsample size & $1000$\\
& $n_{\mathrm{simulation}}$ & Number of subsamples to generate & $1000$\\[6pt]
Bandit-selection & $H$ & Number of iterations & $30$\\
& $b$ & Number of samples to add at each iteration & $10$\\
\hline
\end{tabular}
\end{table}

\begin{figure}[!h]
 \begin{minipage}[b]{.35\textwidth}
  \centering
  \includegraphics[width=1.6in]{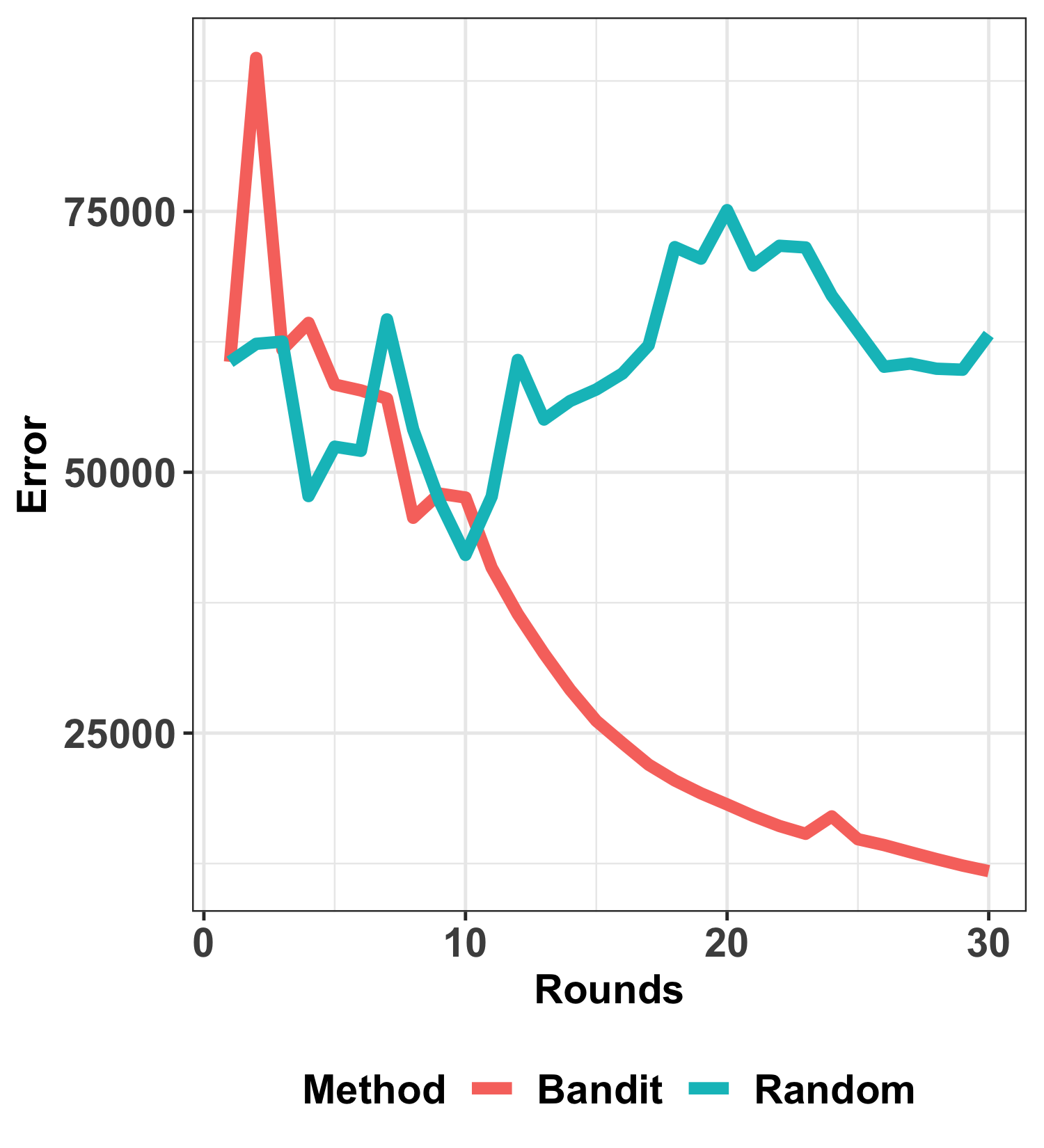}
  \end{minipage}%
  \hfill
  \begin{minipage}[b]{.42\textwidth}
  \centering
  \includegraphics[width=1.9in]{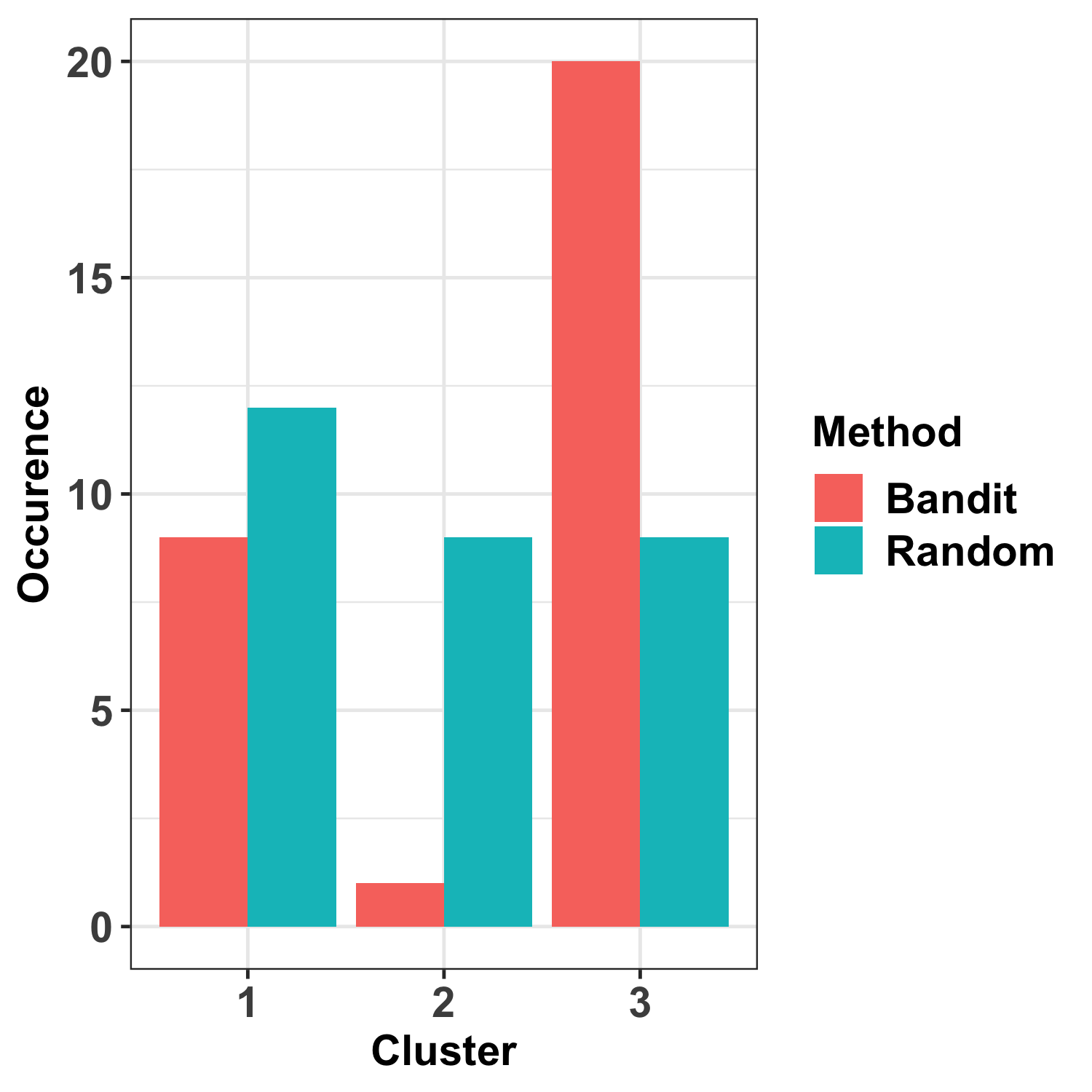}
  \end{minipage} 
  \caption{Bandit selection results for the simulation in section \ref{sec:simulation} when $K=3$ and time $z$ has an effect on the relationship between $y$ and $x$ according to formula \ref{formula:time}. The left panel shows the generalization accuracy over iterations by bandit selection (red) or sequentially adding samples at random (green). The right panel shows the final occurrence of different source subsets correspondingly.}
  \label{fig:time_variant_bandit}
\end{figure}

\begin{figure}
 \begin{minipage}[b]{.38\textwidth}
  \centering
  \includegraphics[width=1.6in]{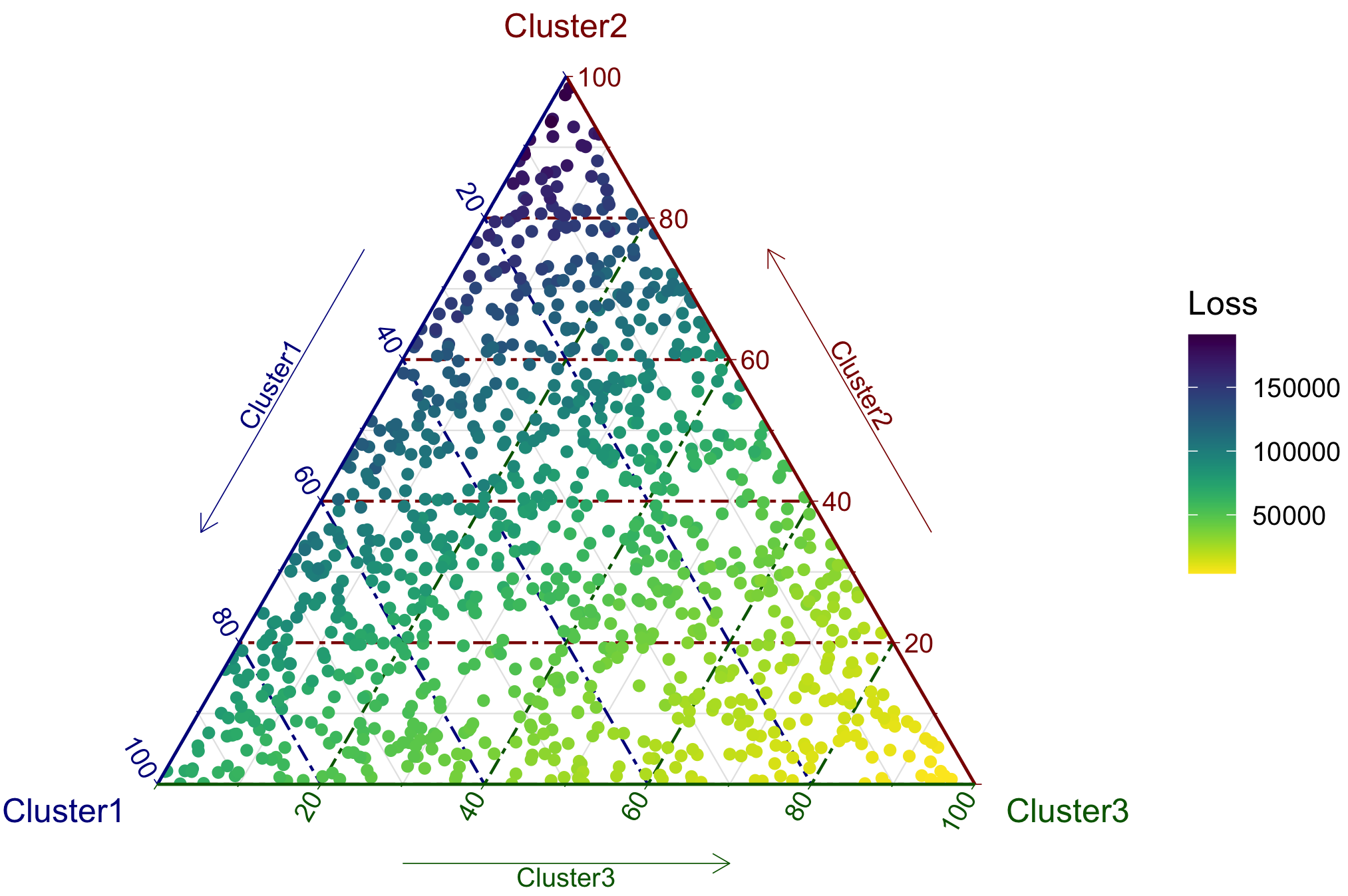}
  \end{minipage}%
  \hfill
  \begin{minipage}[b]{.38\textwidth}
  \centering
  \includegraphics[width=1.6in]{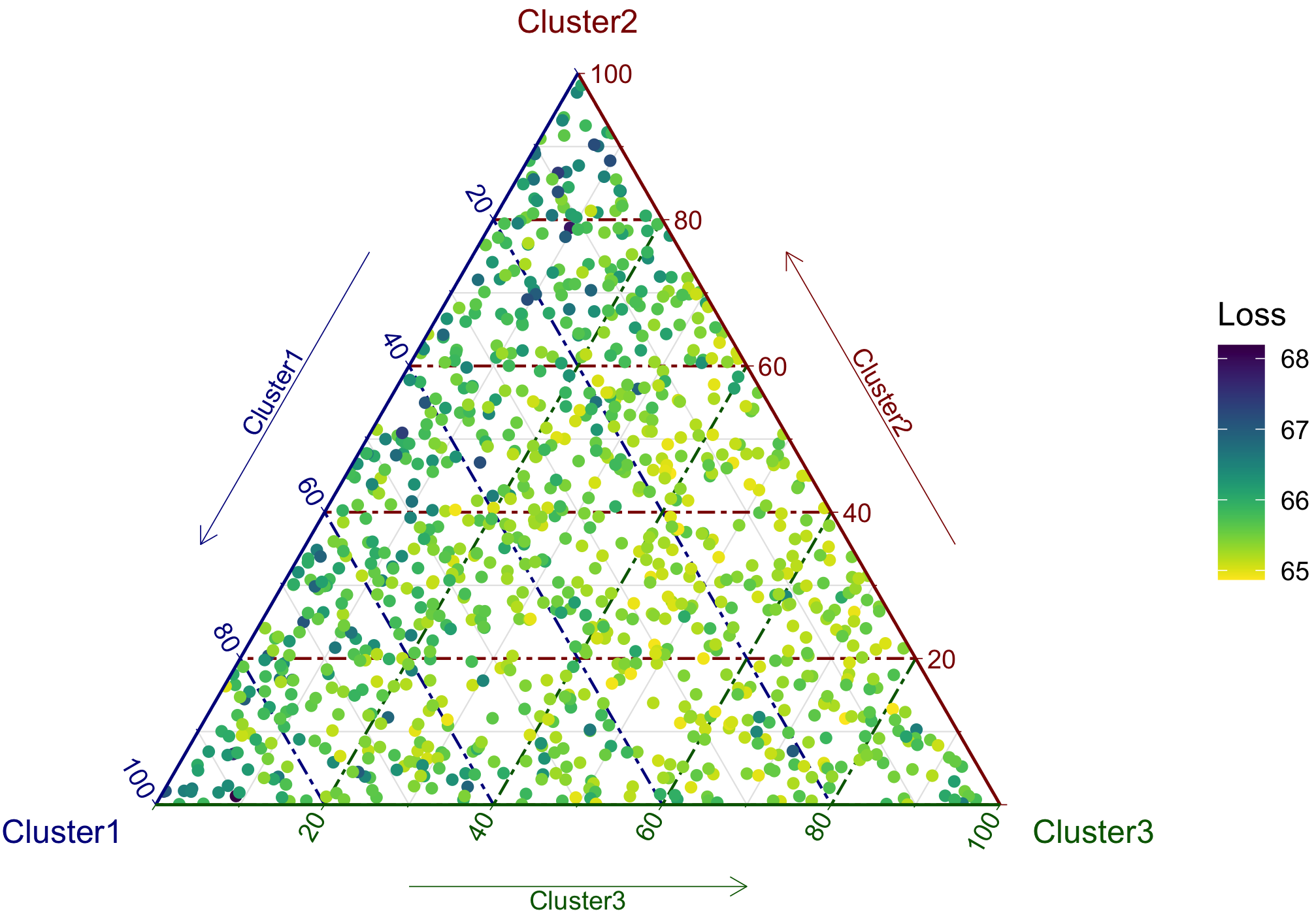}
  \end{minipage} 
  \caption{Generalization performance across different weights of the ensemble method for simulation in section \ref{sec:simulation}. The left panel refers to the case when time has an effect on the relationship between $y$ and $x$ while the right panel refers to the other. In each ternary plot, vertices and edges represent three source subsets and the corresponding axes of weights $w^j$ in subsamples, respectively. Each point inside the triangle indicates a weighted subsample, whose coordinates and color indicate the weights on three source subsets and the prediction loss, respectively.}
  \label{fig:simulation_ensemble}
\end{figure}

Fig. \ref{fig:time_variant_bandit} shows that bandit selection yields a consistently lower loss compared with a random selection, and the performance gap increases over iterations. The final training set of bandit selection mostly comprises data from the latest source subset, which agrees with the intuition that the most recent data are most relevant to the present task. The left panel of Fig. \ref{fig:simulation_ensemble} shows that the ensemble method prefers data from the latest  subset as well. Specifically, higher weights on subsets with large values of $z$ improve performance, reinforcing the conclusions of the bandit method.

\begin{figure}[!h]
 \begin{minipage}[b]{.35\textwidth}
  \centering
  \includegraphics[width=1.6in]{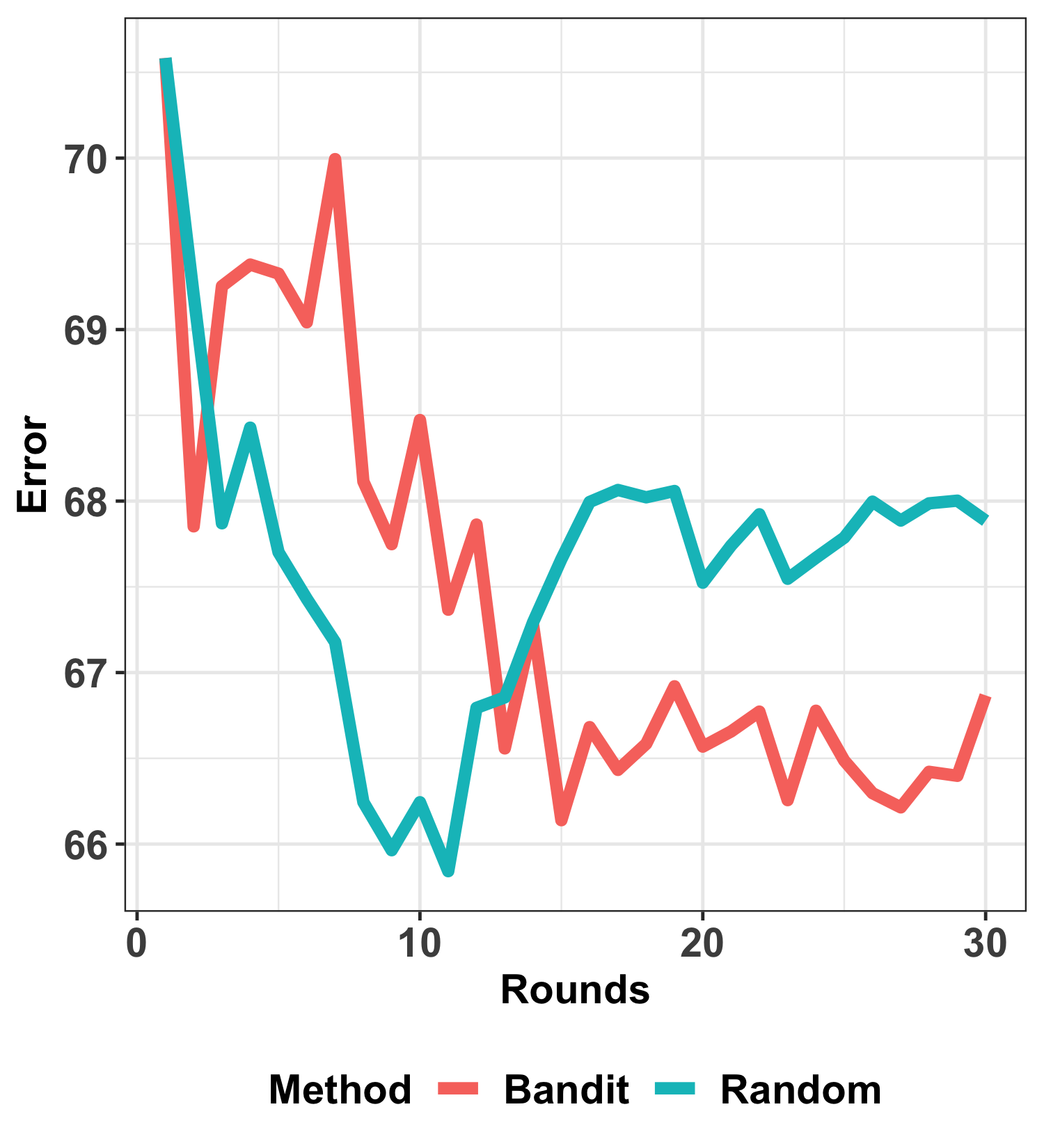}
  \end{minipage}%
  \hfill
  \begin{minipage}[b]{.42\textwidth}
  \centering
  \includegraphics[width=1.9in]{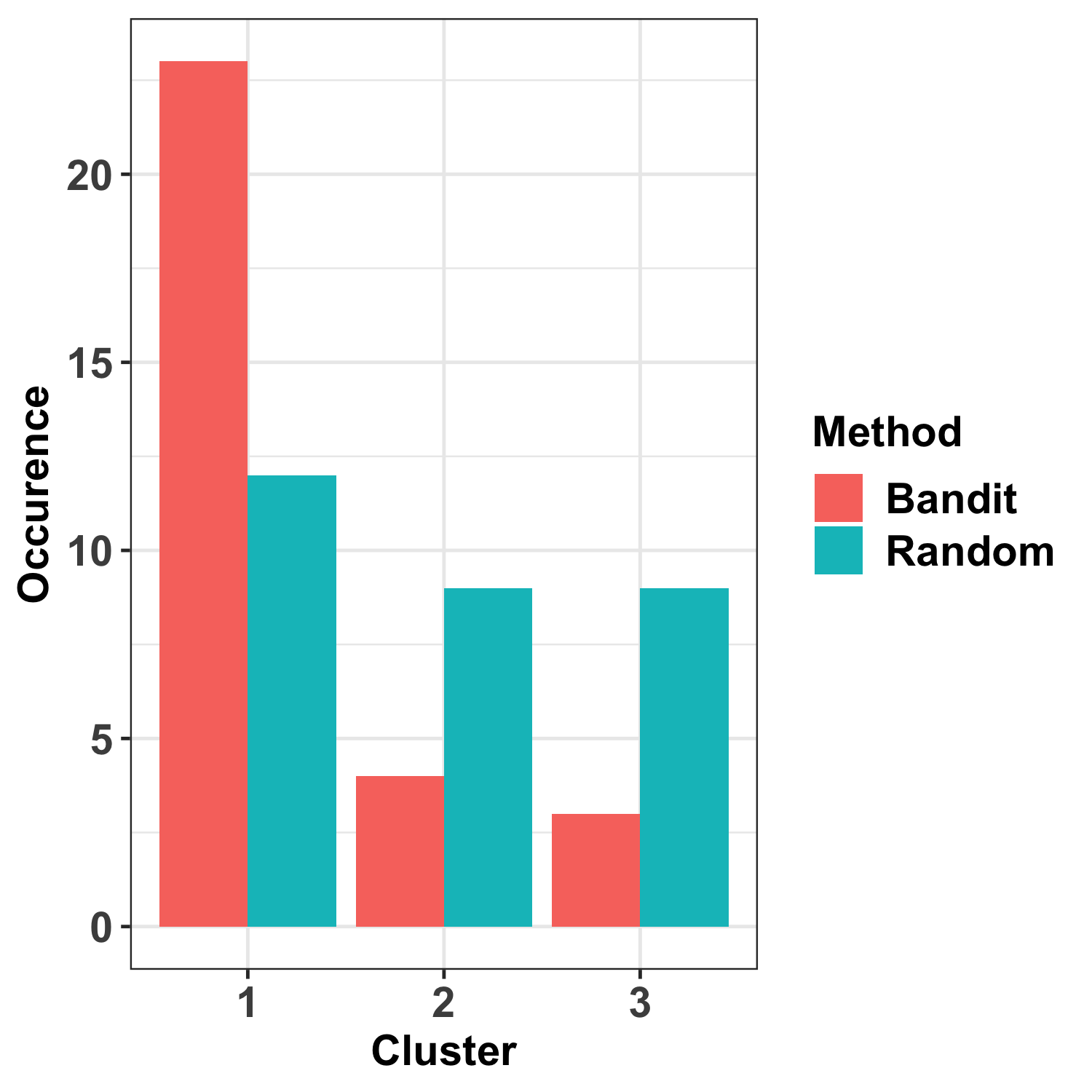}
  \end{minipage} 
  \caption{Bandit selection for simulation in section \ref{sec:simulation} when $K=3$ and time $z$ doesn't have an effect on the relationship between $y$ and $x$. The plots follow the same manners as in Fig. \ref{fig:time_variant_bandit}}
  \label{fig:time_invariant_bandit}
\end{figure}

We repeat these two methods on a dataset where coefficients are constant over time, $y =\beta x+\varepsilon$. Fig. \ref{fig:time_invariant_bandit} and the right panel of Fig. \ref{fig:simulation_ensemble} suggest that neither method improves over a random selection. Under such setting where data are not heterogeneously distributed across contexts (time, in this scenario), a source selection will not be useful. In this sense, our methods can serve as diagnostics on whether an improved source subsample exists.

\section{Experiments}\label{sec:experiment}
\label{sec:data_analysis}
\subsection{California Housing Prices Data}
\label{subsec:california}
\begin{figure}
 \centering
    \includegraphics[scale=0.07]{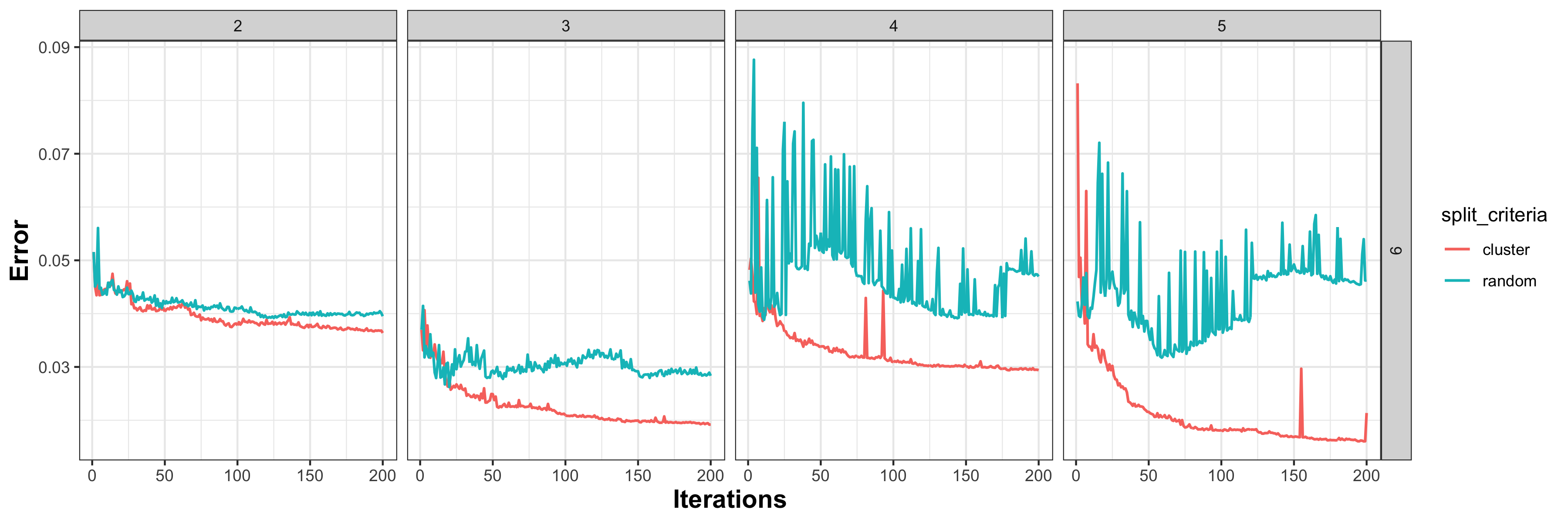}
    \caption{Bandit selection error over iterations with (columns) $K$ from $2$ to $5$ for (red) cluster-based and (green) random source splits in the California Housing Prices experiment (section \ref{subsec:california}). The learned mixture outperforms its alternative consistently with less fluctuations.}
    \label{fig:ca_seed9_K3_lines}
\end{figure}

\begin{figure}
 \centering
    \includegraphics[scale=0.1]{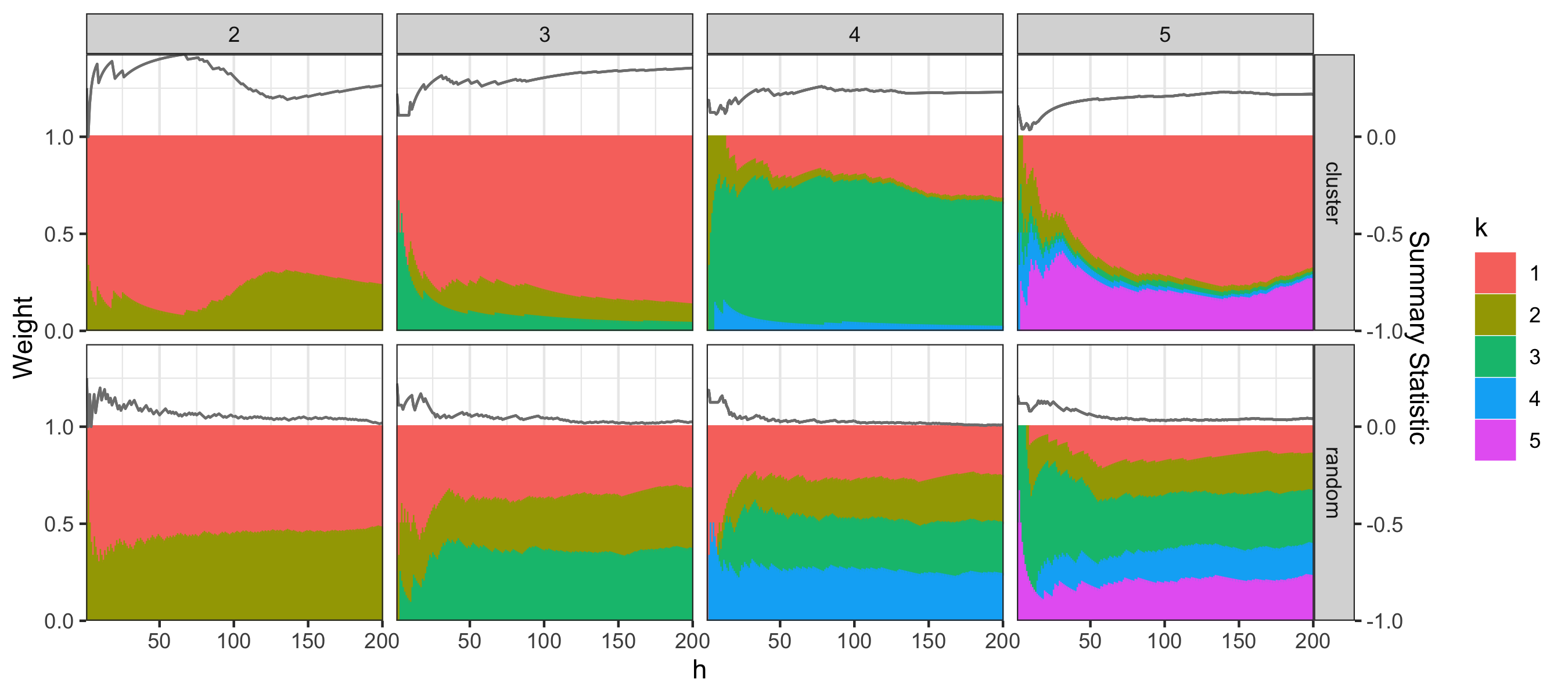}
    \caption{Subset weights and the summary statistic over iterations in the bandit selection of California Housing Prices experiment (section \ref{subsec:california}). Columns refer to different choices of $K$ and rows represent whether the selection is based on (top) multi-armed bandit or (bottom) by random. For each subplot, the x-axis represents iterations, while the y-axis includes two parts: a line plot (top) indicating the summary statistic and stacked bins (bottom) indicating subset weights.}
    \label{fig:ca_seed9_bins}
\end{figure}

\begin{figure}
     \centering
    \includegraphics[scale=0.04]{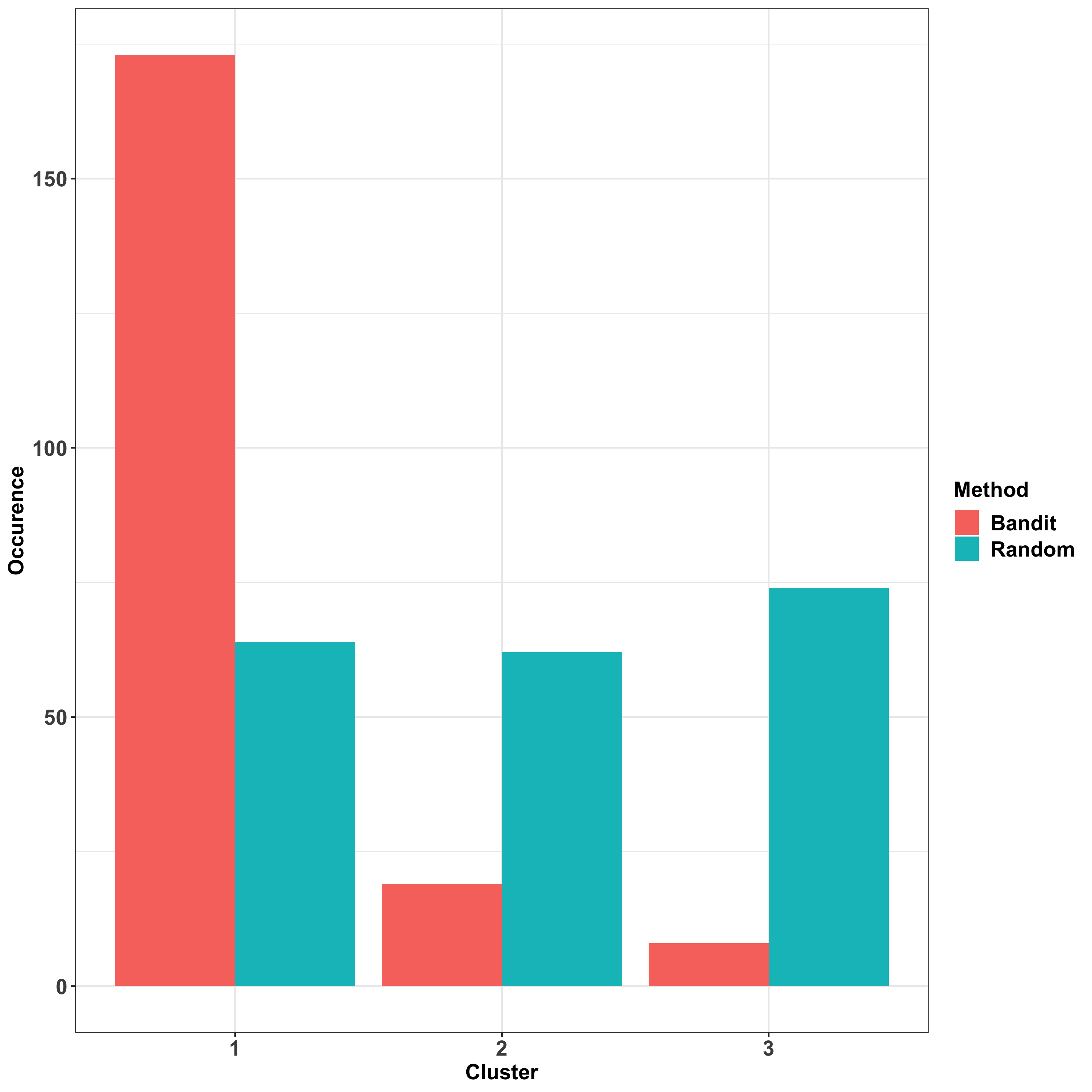}
    \caption{The occurrence of three source subsets over 200 iterations for bandit (red) and random (blue) strategy when $K=3$ in the California Housing Prices experiment (\ref{subsubsec:cal_experiment}). Compared with the random selection which selects almost uniformly across subsets, bandit-based selection selects the first subset $\mathcal{D}_1^s$.}
    \label{fig:ca_seed9_K3_bar}
\end{figure}

\begin{figure}
    \centering
    \includegraphics[scale=0.07]{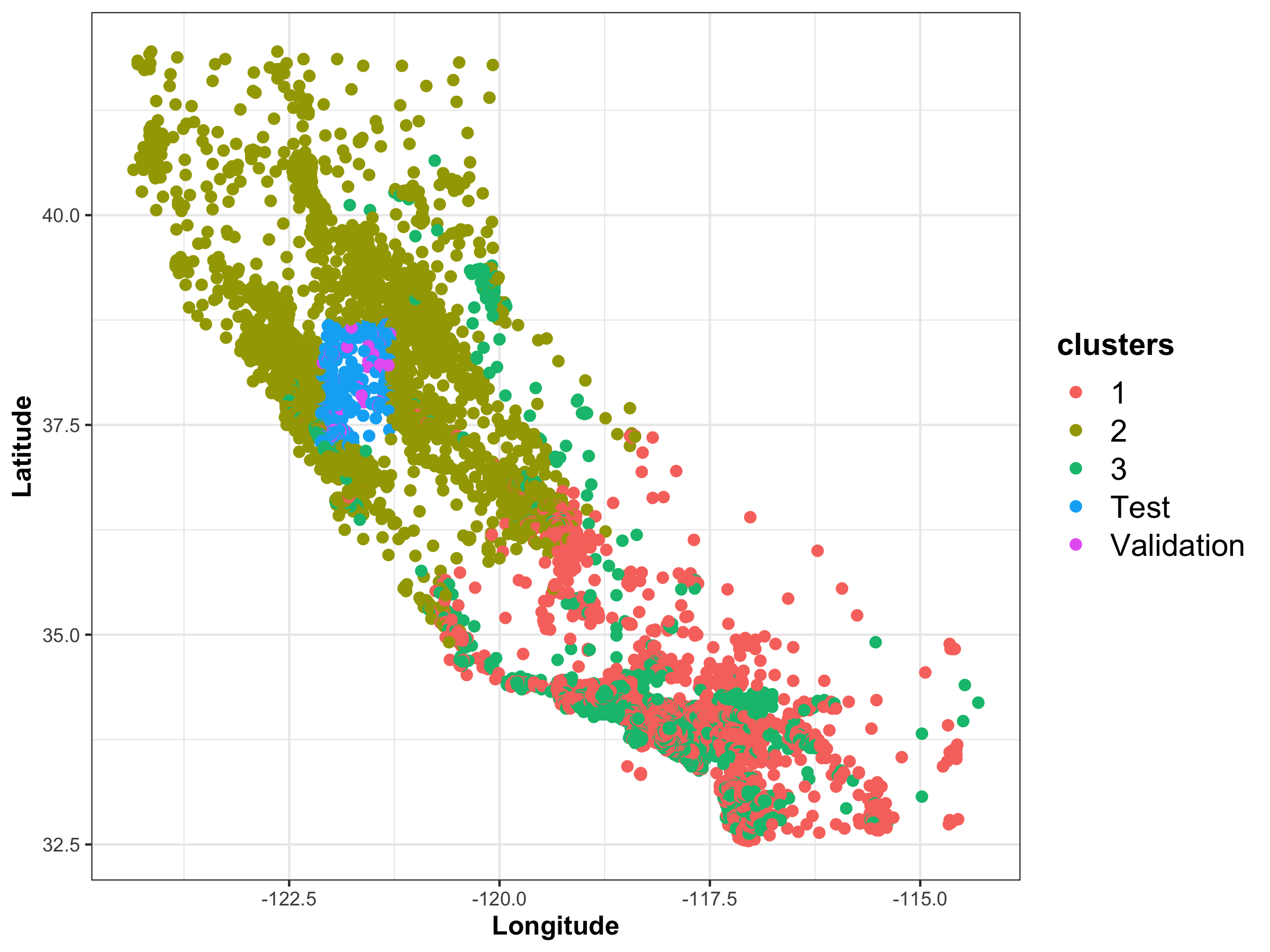}
    \caption{The source-target split on California Housing Prices dataset for $K=3$ with color indicating source subsets or the target dataset (section \ref{subsubsec:cal_experiment}).}
    \label{fig:ca_seed9_K3_split}
\end{figure}

\begin{figure}
    \centering
    \includegraphics[scale=0.1]{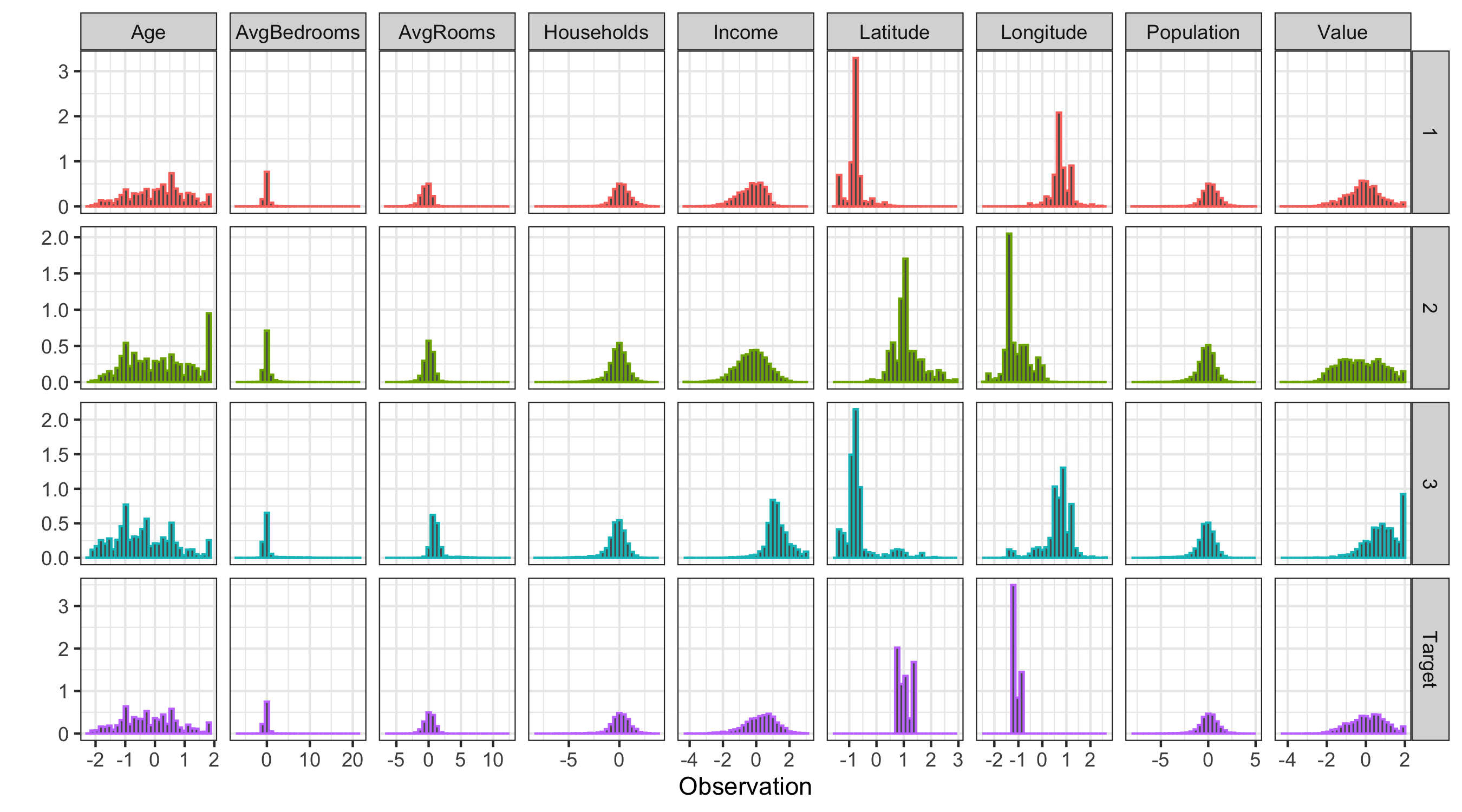}
    \caption{Variable histograms of California Housing Prices dataset (section \ref{subsubsec:cal_experiment}). Rows correspond to source subsets and target, while columns represent different variables. The most favorable source subset, $\mathcal{D}_1^s$, has similar histograms on explanatory variable income and the response variable housing price, whiles its histograms of latitude and longitude show less in common with the target $\mathcal{D}^t$ compared with other source subsets $\mathcal{D}_2^s$ and $\mathcal{D}_3^s$.}
    \label{fig:ca_seed9_K3_variable_hist}
\end{figure}

\begin{figure}
    \centering
    \includegraphics[scale=0.07]{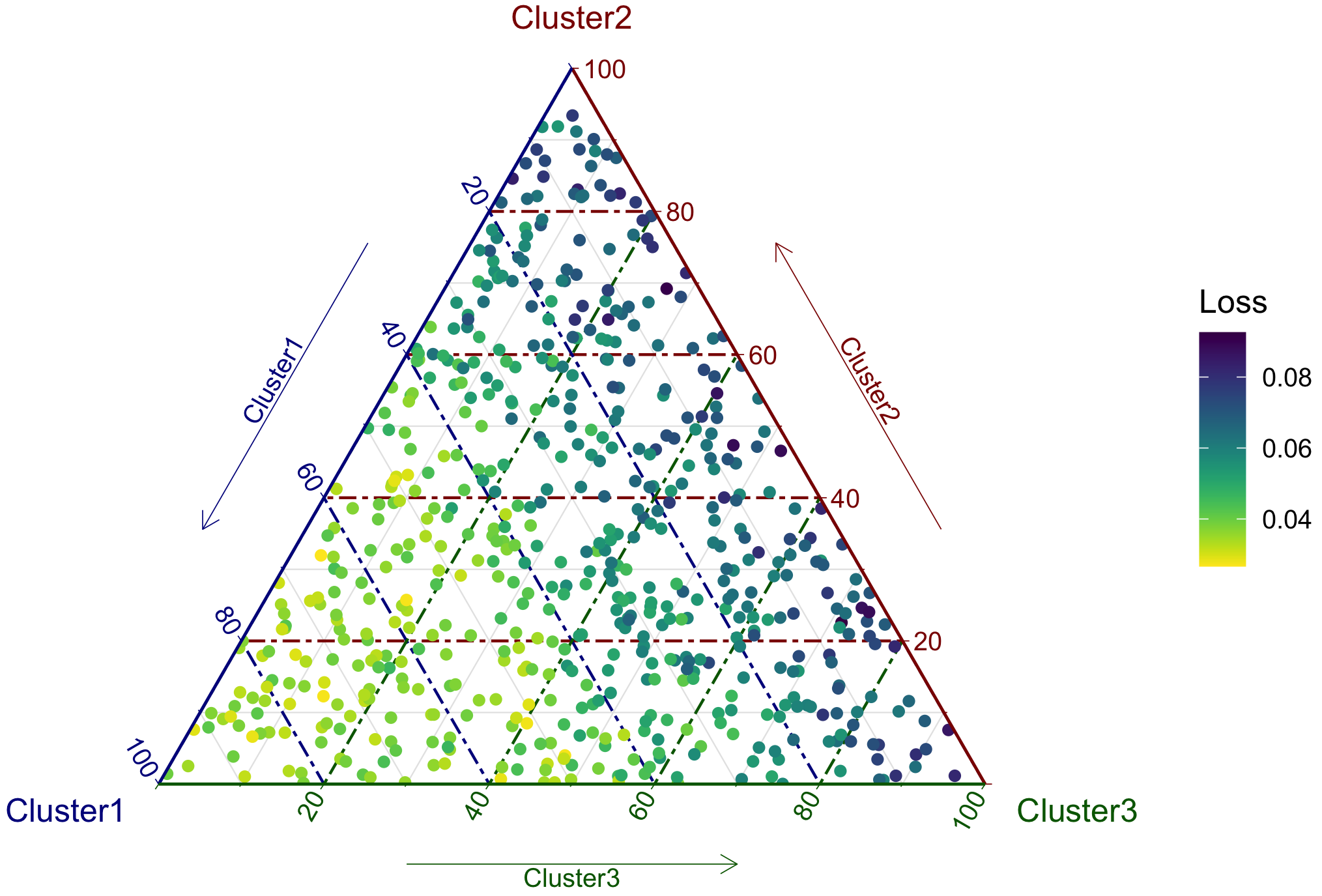}
    \caption{Generalization performance across different weights of ensemble method when $K=3$ in the California housing experiment (section \ref{subsec:california}). The ternary plot follows the same format with Fig. \ref{fig:simulation_ensemble}. Since points near the $\mathcal{D}_1^s$ vertex have lowest loss values, we can draw the same conclusion as drawn from Fig. \ref{fig:ca_seed9_K3_bar} that $\mathcal{D}_1^s$ is more informative.}
    \label{fig:ca_seed9_K3_tern}
\end{figure}

The California Housing Prices Dataset contains median house prices and ten explanatory variables derived from the 1990 census \citep{pace1997sparse}. The data have a spatial component, with each home associated with its latitude and longitude. 
We use this data to explore the scenario where source and target samples share the same variables but are geographically dissimilar, and where a careful selection of source data may support improvement on a target task.
To mimic the situation where the goal is to perform well in a specific region, we geographically split data into source and target. We first fix a range of longitude and latitude to be the target region, and then split the remaining (source) samples via a $K$-means clustering over all variables. The training model is linear regression. Detailed parameters are listed in Table \ref{tab:ca_para} and results are given through Figs. \ref{fig:ca_seed9_K3_lines} -  \ref{fig:ca_seed9_K3_tern}. 
The high-level takeaways are
\begin{itemize}
    \item The bandit selection method results in a source subsample that consistently improves performance relative to a random selection.
    \item The final bandit selection prefers certain source subsets over others, and the summary statistic has a high value.
    \item The ensemble and bandit selection methods agree with each other on the selected source subsets. 
\end{itemize}

Fig. \ref{fig:ca_seed9_K3_lines} suggests that bandit selection yields consistently better and more robust performance than the random selection for $K \in \{2, \dots, 5\}$. The two error curves have increasing gaps after overlapping in the first few iterations. After 100 iterations, the bandit selection error curve consistently has lower target set error than random selection. From Fig. \ref{fig:ca_seed9_bins}, the increasing summary statistic in bandit selection reflects the increasing nonuniformity of weights across source subsets. The selection makes certain subsets more visible, with individual subsets contributing more than 50\% of the total weight. In comparison, the random selection shows decreasing summary statistics and finally results in almost uniform weights over the source subsets. Further, the bandit selection's preference on source subsets also agrees with the ensemble search; \textit{e.g.} from Fig. \ref{fig:ca_seed9_K3_bar} and \ref{fig:ca_seed9_K3_tern} both methods favor the first source subset when $K=3$. Feature histograms for each source subset and the target (Fig. \ref{fig:ca_seed9_K3_variable_hist}) suggests that this preferred subset $\mathcal{D}_1^s$ is more similar with the target $\mathcal{D}^t$ with respect to the explanatory variable income and the response variable housing price. Those two variables may be responsible for the selection of specific source clusters for this given target.
\begin{table}[!h]
\caption{Model parameters of the California housing dataset}
\label{tab:ca_para}
\begin{tabular}{@{}llll@{}}
\hline
Sampling Procedure & Notation & Description & Value \\
\hline
{Both methods} & $K$ & Number of clusters in source & $2, 3, 4, 5$ \\[6pt]
{Ensemble method} & $n_{\mathrm{training}}$ & Subsample size & $200$\\
& $n_{\mathrm{simulation}}$ & Number of subsamples to generate & $200$\\[6pt]
Bandit-selection & $H$ & Number of iterations & $200$\\
& $b$ & Number of samples to add at each iteration & $20$\\
\hline
\end{tabular}
\end{table}

\subsection{Camelyon17-wilds dataset}
\label{subsec:camelyon17}
In breast cancer diagnosis, pathologists detect tumors manually by screening slides of lymph nodes under microscope. While automated algorithms generally yield good results, they may generalize poorly across facilities or equipment due to differences in data collection and processing \citep{tellez2019quantifying}. We study the tumor classification task on Camelyon17-wilds dataset, where inputs are $96 \times 96$ patches of whole-slide images of sentinel lymph nodes, and outputs are their binary tumor indicator labels for breast cancer metastases detection \citep{koh2020wilds}. The data consist of $455,955$ patches collected from five separate hospitals, across which generalization performances differ substantially. 

The Camelyon17-wilds dataset distinguishes itself from the simulation (section \ref{sec:simulation}) and California Housing Prices experiment (section \ref{subsec:california}) in two aspects. First, it consists of images whose original features, pixels, are only weakly predictive; this makes extracting higher level representations necessary. 
Since deep learning approaches are usually used for imagery data, the model parameters are not directly interpretable with respect to input features. 
High-dimensionality of feature representations also requires dimensionality reduction methods be employed before reaching conclusions. Second, data are collected from five hospitals, which can be treated as inherent group splits with potential distributional shifts. As discussed by \cite{koh2020wilds}, generalization from one hospital to another may have bad performance. This makes source data selection realistic and sensible. 

The experimental protocol follows from section \ref{subsec:protocols}. The classification model we use is the Residual Network (ResNet) proposed by \cite{he2016deep} with implementation details described in Appendix \ref{subsec:appendix_cal_model}. For the source / target split, we set one hospital as the target. For the source, we use either samples from all but the target hospital or samples from all hospitals (including the target hospital). Notice that for the second arrangement, the source and target still don't contain overlapping data points, although some observations in the source come from the same hospital as the target.
The source data are further split by either (i) hospitals themselves or (ii) clustered deep neural network representations of the imagery. By changing the target hospital, we can also evaluate whether samples of certain hospitals are ``uniformly easier'' to predict, and whether the good generalization is commutative between hospitals. 

We provide details of the source splitting approaches.
Approach (i) uses the original hospital IDs as source subset splits ($K=5$) to mimic the situation when we select source data from facilities available. We also assign subsets randomly as a comparison.
In contrast, approach (ii) is based on extracting features from a ResNet. Specifically, we pass a pretrained ResNet through source images and obtain activations after convolution layers. Each convolution layer corresponds to an array of activations with first dimension equal to the number of data points. 
After flattening the activation array and performing Principal Component Analysis with $50$ components, we have a $455,955 \times 50$ matrix of feature representations from each convolution layer. 
We use activations from either shallow or deep convolutional layers to study the influence of this choice. In principle, representations from deeper layers in the network capture higher level structures. However, it may reveal too much about the class label and thus cannot be a good criteria to split the source images. For example, if the deepest activations are discriminative enough for classification, then splitting the source accordingly can result in highly imbalanced subsets with respect to labels.
With either deep or shallow ResNet feature representations, we cluster the source into $K$ subsets via $K$-means.

\subsubsection{Experimental results}
\label{subsubsec:cal_experiment}
\begin{figure}[htp]
    \centering
    \includegraphics[scale=0.06]{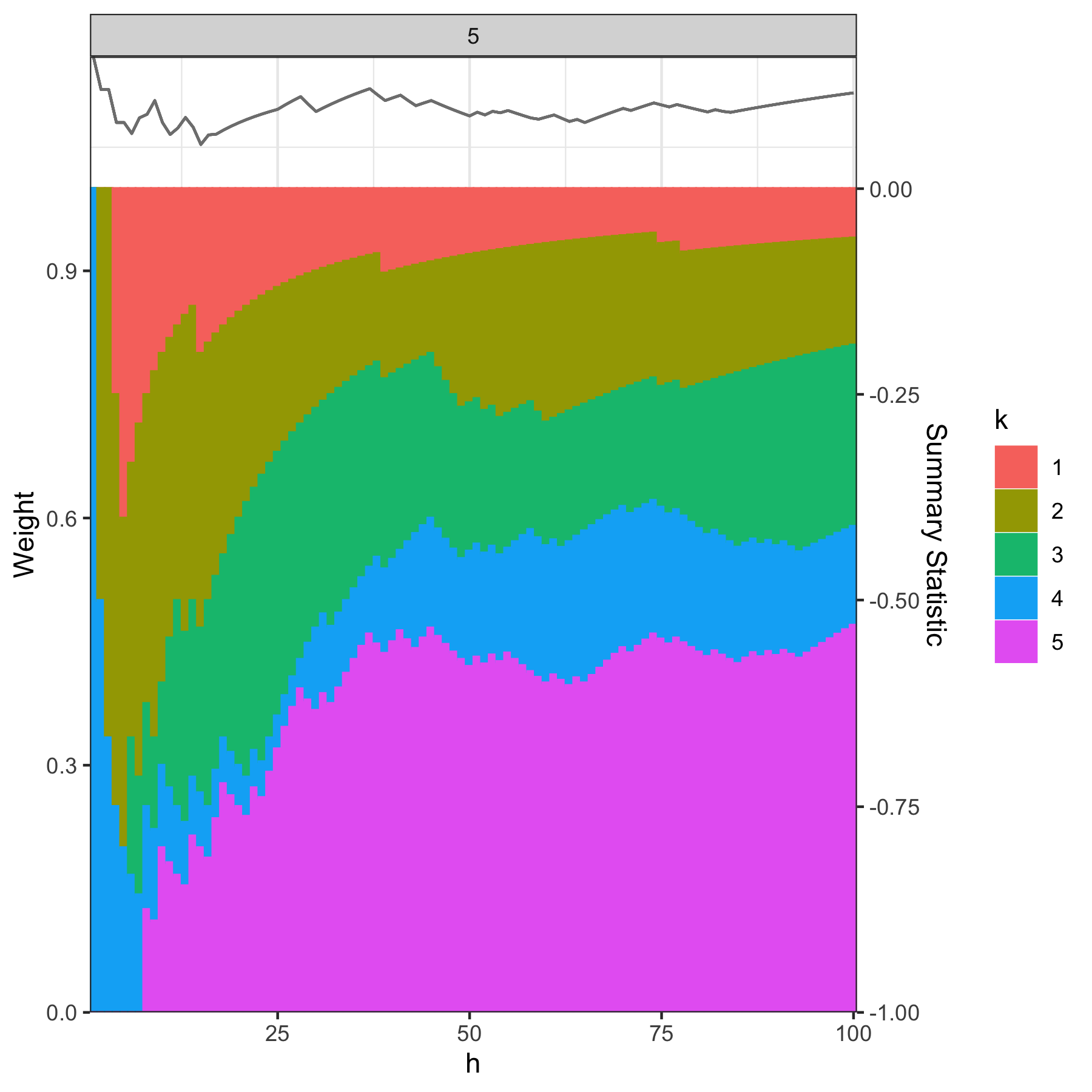}
    \caption{Hospital-based subset weights as a function of bandit selection iteration. Target data consists of hospital 5 only, while source consists of observations from all hospitals. We split the source into $K=5$ subsets according to hospitals. The plot follows the same format as each subplot of Fig. \ref{fig:ca_seed9_bins}.}
    \label{fig:all_stack_center4}
\end{figure}

\begin{figure}[htp]
    \centering
    \includegraphics[scale=0.09]{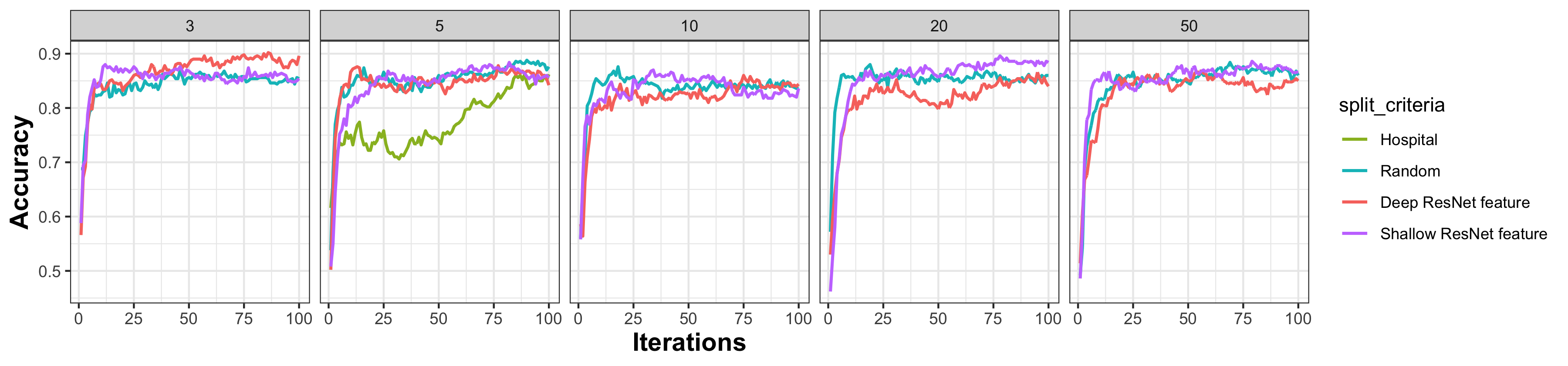}
    \caption{Model accuracy over bandit selection. Target data consists of hospital 5 observations only, while source consists of observations from all hospitals. The source is split into $K$ subsets either at random, by hospital ID (only when $K=5$), by clustering the deep ResNet representations, or by clustering the shallow ResNet representations.  Column represents number of source subsets $K$ and in each subplot, $x$ and $y$-axes represent iterations and accuracy, respectively, with color indicating the ways we split the source. In each bandit selection, the model converges within the first fifteen iterations maintains accuracy between $0.8$ and $0.9$ with small fluctuations. The result doesn't have an obvious dependence on $K$ or how we split the source.}
    \label{fig:all_lines_target_4}
\end{figure}

\begin{figure}[h]
    \centering
    \includegraphics[scale=0.09]{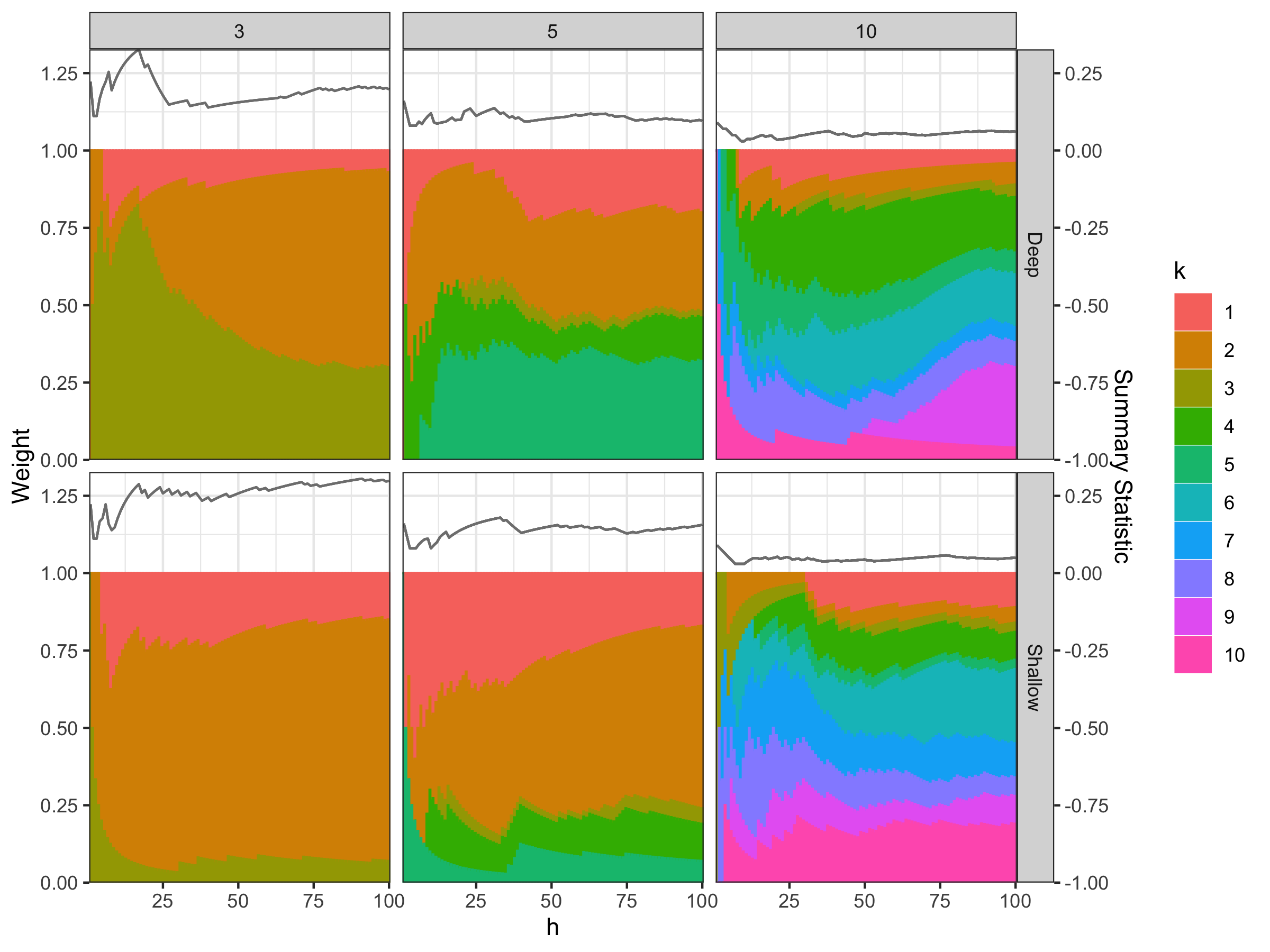}
    \caption{Hospital-based subset weights as a function of bandit selection iteration in the Camelyon17-wilds experiment (section \ref{subsubsec:cal_experiment}). Target data consists of hospital 5 only, while source consists of observations from all hospitals. We cluster the source into $K$ subsets ($K$ indicated by column) according to deep (top) or shallow (bottom) ResNet representations. Each subplot follows the same format as each subplot of Fig. \ref{fig:ca_seed9_bins}.}
    \label{fig:all_stack_cluster4}
\end{figure}

We start with the case where the target is hospital 5 and the source contains data from all hospitals. First, we investigate the influence of source splitting criteria on target performance. When we use hospitals as source subsets ($K=5$), we would expect to select most samples from hospital 5, since variation among facilities may affect generalization. Fig. \ref{fig:all_stack_center4} confirms this conjecture under the bandit selection, since the final source composition is dominated by hospital 5. Deciding source subsets by clustering ResNet features yields a comparable accuracy over bandit iterations (Fig. \ref{fig:all_lines_target_4}) -- accuracy on the target rises in the first 10 iterations and stabilizes afterwards. The choice of $K$ does not noticeably affect accuracy, but does affect the distribution of source subsets used. When $K = 5$, certain clusters are preferred and almost dominate the selection (Fig. \ref{fig:all_stack_cluster4}). The summary statistic reaches a similar value as the hospital split case. A larger choice of $K$ results in a more uniform spread of the ultimate subset weights as well as a lower summary statistic. This is expected since finer clustering makes clusters less distinguishable from one other. 

\begin{figure}[h]
    \centering
    \includegraphics[scale=0.09]{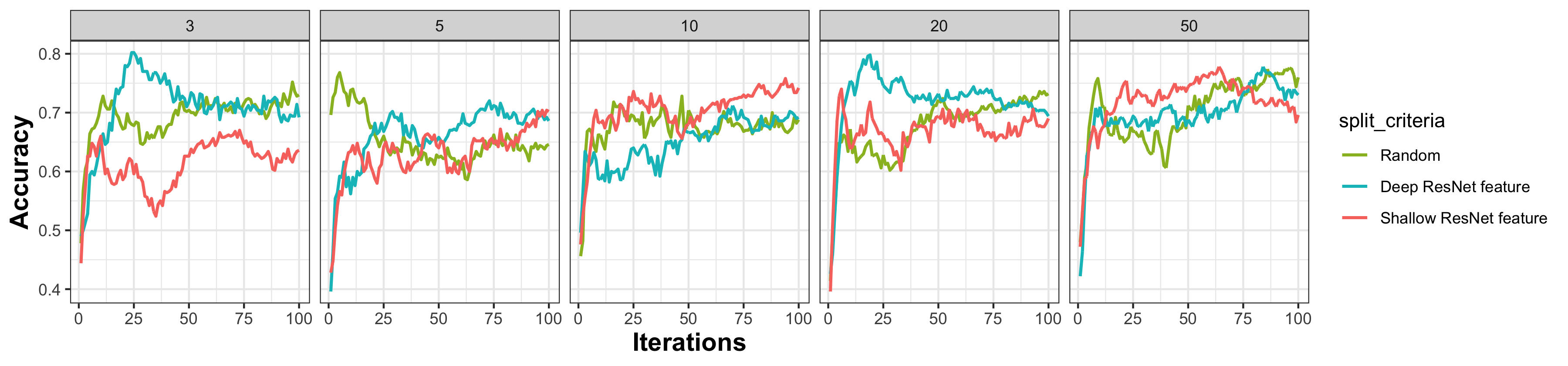}
    \caption{Model accuracy over bandit selection iterations in the Camelyon17-wilds experiment for generalization to hospital 5(section \ref{subsubsec:cal_experiment}). The experimental settings are the same as Fig. \ref{fig:all_lines_target_4} except that data from hospital 5 are excluded from the source. In each bandit selection, model accuracy stabilizes between $0.6$ and $0.75$. Performance has a positive relationship with $K$, but does not have a clear dependence on split criteria.}
    \label{fig:other_lines_target_4}
\end{figure}

Next, we change the source by excluding data points from hospital 5. Compared with previous results, the accuracy of bandit selection fluctuates much more after consistently increasing in the first 10 iterations, and it converges to a lower overall accuracy (Fig. \ref{fig:other_lines_target_4}). This further confirms that including data from hospital 5 in the source supports generalization. Another observation is that in both source settings, splitting the source randomly yields a comparable bandit selection. The source split by clustering deep neural network features does not ensure a better source subsample. We discuss this phenomenon further in section \ref{subsubsec:transition}.

We have focused on results where the target consists of hospital 5 samples. Other choices of target hospitals yield similar results. The only noticeable difference is that, when we use hospital 1 as the target and all hospitals as source and split the source by hospitals, the bandit selection does not prefer data from hospital 1 over other hospitals. 
Complete experimental results can be found in Appendix \ref{subsec:appendix_cal_experiment_results}. If we treat our method as a diagnostic, we would conclude that prediction on hospital 1 does not require source selection. 
This agrees with \citep{koh2020wilds}, which frames hospital 5 as the out-of-distribution data and other hospitals as in-distribution data. In the following subsection, we explore to what extent having hospital 5 data in the source benefits generalization.



\subsubsection{Transitional study}
\label{subsubsec:transition}
Having source samples from hospital 5 benefits generalization performance. A follow-up question is the extent to which improvement depends on the proportion of source data points available from hospital 5. For example, are a few samples from hospital 5 sufficient to guarantee strong generalization? This relates to the situation where we must determine the number of samples from the hospital of interest that will need to be labeled, assuming plentiful data from other hospitals. Such knowledge is especially important when labeling is scarce and costly.

In the following experiment, we perform bandit selection with the source containing $1500$ examples from hospitals 1 through 4 each and varying numbers of examples from hospital 5. We call the size of hospital 5 data in the source compared with others the ``proportion'' parameter. When the proportion equals to 0, then we do not have data from hospital 5 in the source; when the proportion equals to 1, then all 1500 source examples are from hospital 5. We initialize the model by training a ResNet classifier with data from hospital 1 to 4 in the source until its training accuracy reaches $0.7$. That mimics the scenario where a model must be adapted to a target facility, after having been initialized using others. At each bandit iteration, we update the model with one selected source subset for 5 epochs. For source split criteria, we continue to use hospitals, clusters, and random assignments.

\begin{figure}[htp]
    \centering
    \includegraphics[scale=0.1]{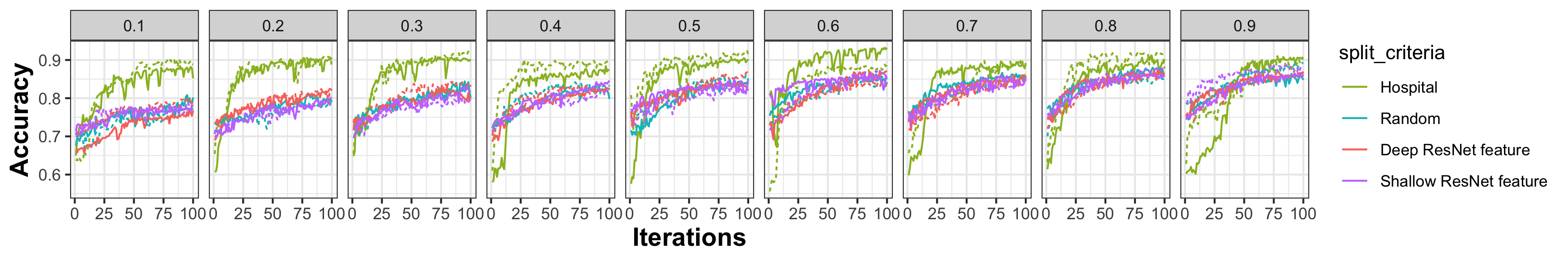}
    \caption{Transitional study: generalization accuracy over iterations in bandit selection with ``proportion'' parameter denoted by columns (section \ref{subsubsec:transition}). The format of each subplot is the same as subplots in Fig. \ref{fig:all_lines_target_4}. }
    \label{fig:transition_line}
\end{figure}

Fig. \ref{fig:transition_line} suggests that adding hospital 5 data points benefits cluster-based bandit selection and random selection almost equally. In contrast, the hospital-based selection starts with a lower performance at first, increases faster and outperforms them within ten iterations. There is a distinct gap between the performance of hospital-based selection and the rest. However, hospital-based selection seems to benefit little from adding further samples from hospital 5, and its advantage over others becomes less obvious when the source includes as many data points from hospital 5 as data from any other hospital.

\begin{figure}[htp]
    \centering
    \includegraphics[scale=0.08]{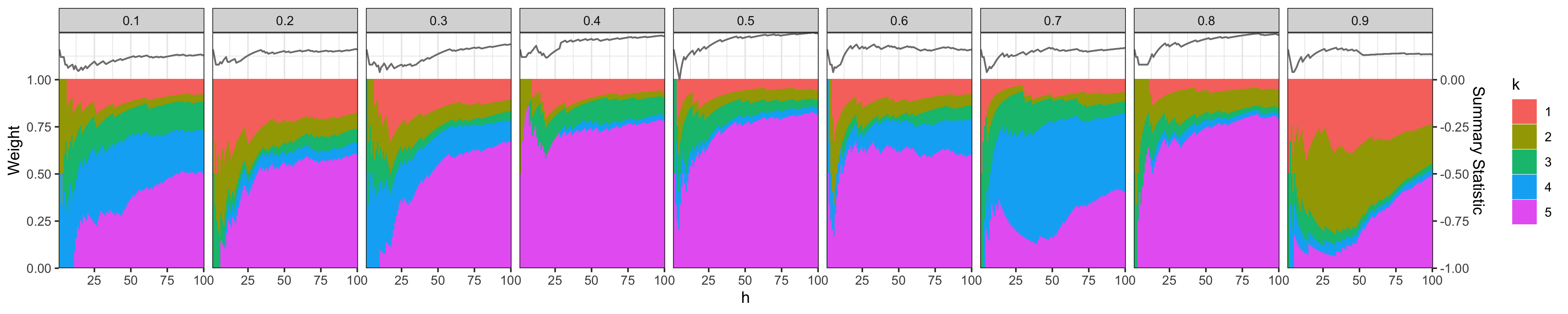}
    \caption{Hospital-based subset weights as iteration goes by in bandit selection of transitional study (section \ref{subsubsec:transition}). Target data consists of hospital 5 only, while source consists of observations from all hospitals. Columns refer to the choices of ``proportion'' parameter and subplots follow the same manner as Fig. \ref{fig:all_stack_center4}.}
    \label{fig:transition_stack_center}
\end{figure}

Fig. \ref{fig:transition_stack_center} explains the behavior of hospital-based selection. Good overall performance results from the fact that hospital 5 is eventually selected with high probability. Unstable initial performance comes from the randomness of initialization in bandit selection -- by chance, the bandit may initially place low weight on hospital 5. In contrast, in the other two scenarios, the training set always contains data from hospital 5, regardless of the choices from the bandit selection. This ensures a relatively high starting performance. Since the bandit selection allows training examples being repeatedly chosen, increasing the number of data points from hospital 5 in the source does not necessarily benefit performance. In fact, performance even drops slightly when the proportion is close to $1$. An intuition behind this is, when we have few samples from hospital 5, the algorithm tends to repeatedly select and train on the same data, while it tends to train on a variety of data points from hospital 5 otherwise. In this sense, we don't necessarily need as many samples from the source. 

We make two secondary comments on Fig. \ref{fig:transition_line}. First, similar behavior of two repetitions suggests the robustness of the algorithm. Second, clustering according to either deep or shallow ResNet feature yields similar results. This contradicts the intuition that deep features may capture dissimilarity across facilities. In this problem, in order to have good generalization on hospital 5, we need to include (a few) data points from hospital 5, and allow weight to concentrate on that source.


\section{Conclusion}
We have studied two source selection methods in out-of-distribution generalization problems. We demonstrate how our methods work in one simulated and two real-life datasets, discussing reasons why selection may or may not help across specific contexts. Both methods can also serve as diagnostics and provide quantitative and qualitative illustrations of dataset heterogeneity. One direction for further study is further investigation of criteria for source splitting. For example, the choice of data representations and number of source subsets would benefit from further analysis.

\section{Acknowledgement}
This research was performed using the compute resources and assistance of the UW-Madison Center For High Throughput Computing (CHTC) in the Department of Computer Sciences. The CHTC is supported by UW-Madison, the Advanced Computing Initiative, the Wisconsin Alumni Research Foundation, the Wisconsin Institutes for Discovery, and the National Science Foundation, and is an active member of the OSG Consortium, which is supported by the National Science Foundation and the U.S. Department of Energy's Office of Science.


\begin{appendices}
\section{Supplementary materials of experiments on Camelyon 17-wilds datset}
\subsection{Model}
\label{subsec:appendix_cal_model}

Our training model is the ResNet model \citep{he2016deep} with 18 layers pretrained by PyTorch (resnet18). In the experiments. we randomly select $5000$ data points evenly distributed over hospitals and labels. The batch size is $16$. In the bandit selection of section \ref{subsubsec:cal_experiment}, we add $b=30$ data points from the chosen source subset to the training data and train the whole model for $30$ epochs at each of the $H=100$ iterations. In the bandit selection of section \ref{subsubsec:transition}, we initiate with the model trained on data points from hospitals 1 to 4 with training accuracy $0.70$ (it hasn't reached its capacity). At each of the $H=100$ iterations, we add $b=100$ samples from one source subset into the training set and train the model for $3$ epochs.

In the case that we cluster the source via features, we extract the features of the first and last convolutional layers by passing the source data (image arrays) through the ResNet model we mentioned earlier. After that, we reduce the dimension of either case into $50$ using PCA. For simplicity, we denote the resulting arrays of the first layer as shallow features and that of the last layer as called deep features.

\subsection{Experimental results}
\label{subsec:appendix_cal_experiment_results}

\begin{figure}[htp]
    \centering
    \includegraphics[scale=0.07]{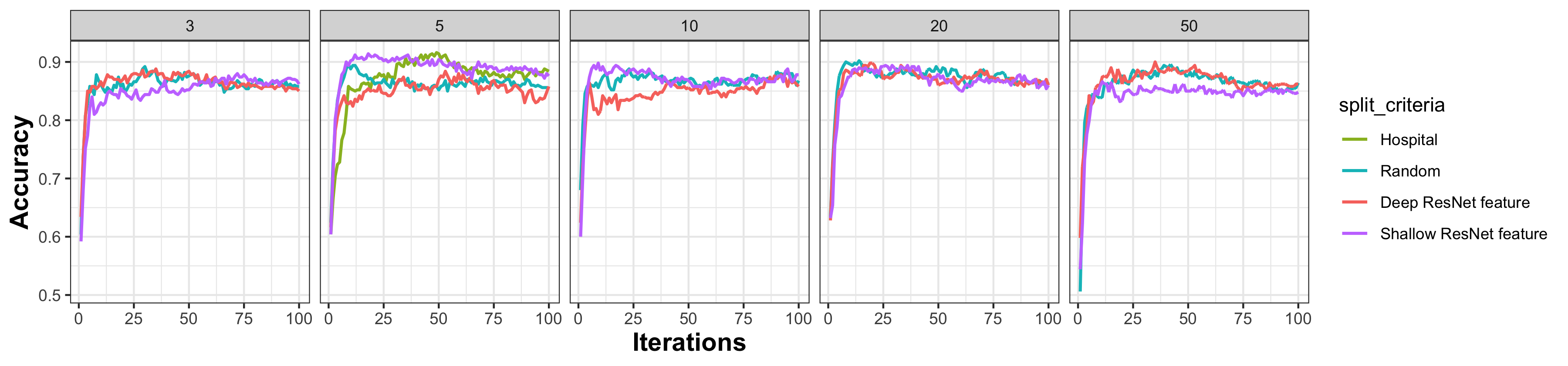}    
    \caption{Model accuracy over bandit selection. Target data consists of hospital 1 observations only, while source consist of observations from all hospitals. The experiment and visualization settings are the same as Fig. \ref{fig:all_lines_target_4} except that the target hospital is hospital 1.}
    \label{fig:all_lines_target_0}
\end{figure}

\begin{figure}[htp]
    \centering
    \includegraphics[scale=0.08]{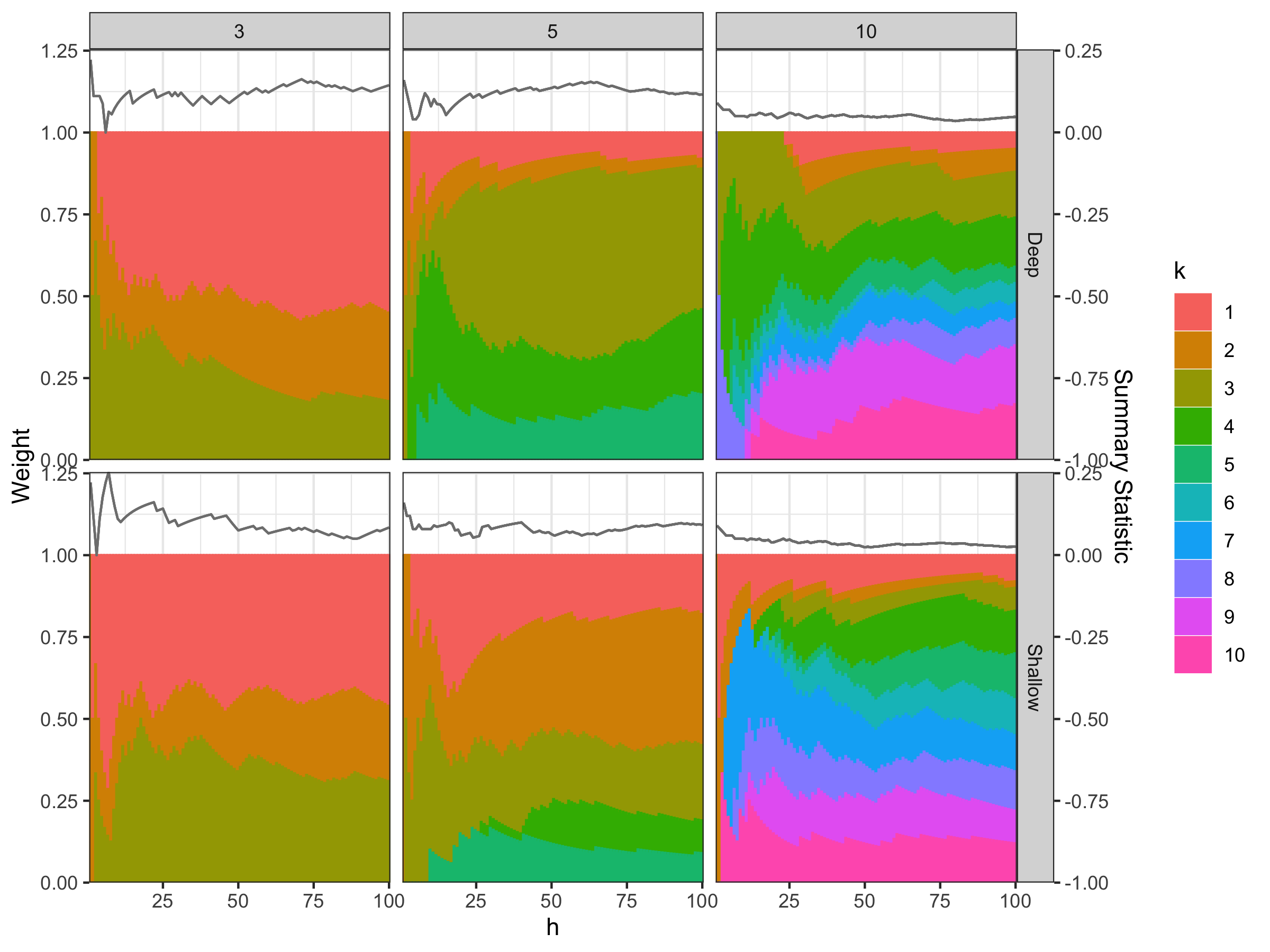}
    \caption{Hospital-based subset weights as iteration goes by in bandit selection. The experiment and visualization settings are the same as Fig. \ref{fig:all_stack_cluster4} except that the target hospital is hospital 1.}
    \label{fig:all_stack_cluster0}
\end{figure}

\begin{figure}[!htp]
    \centering
    \includegraphics[scale=0.08]{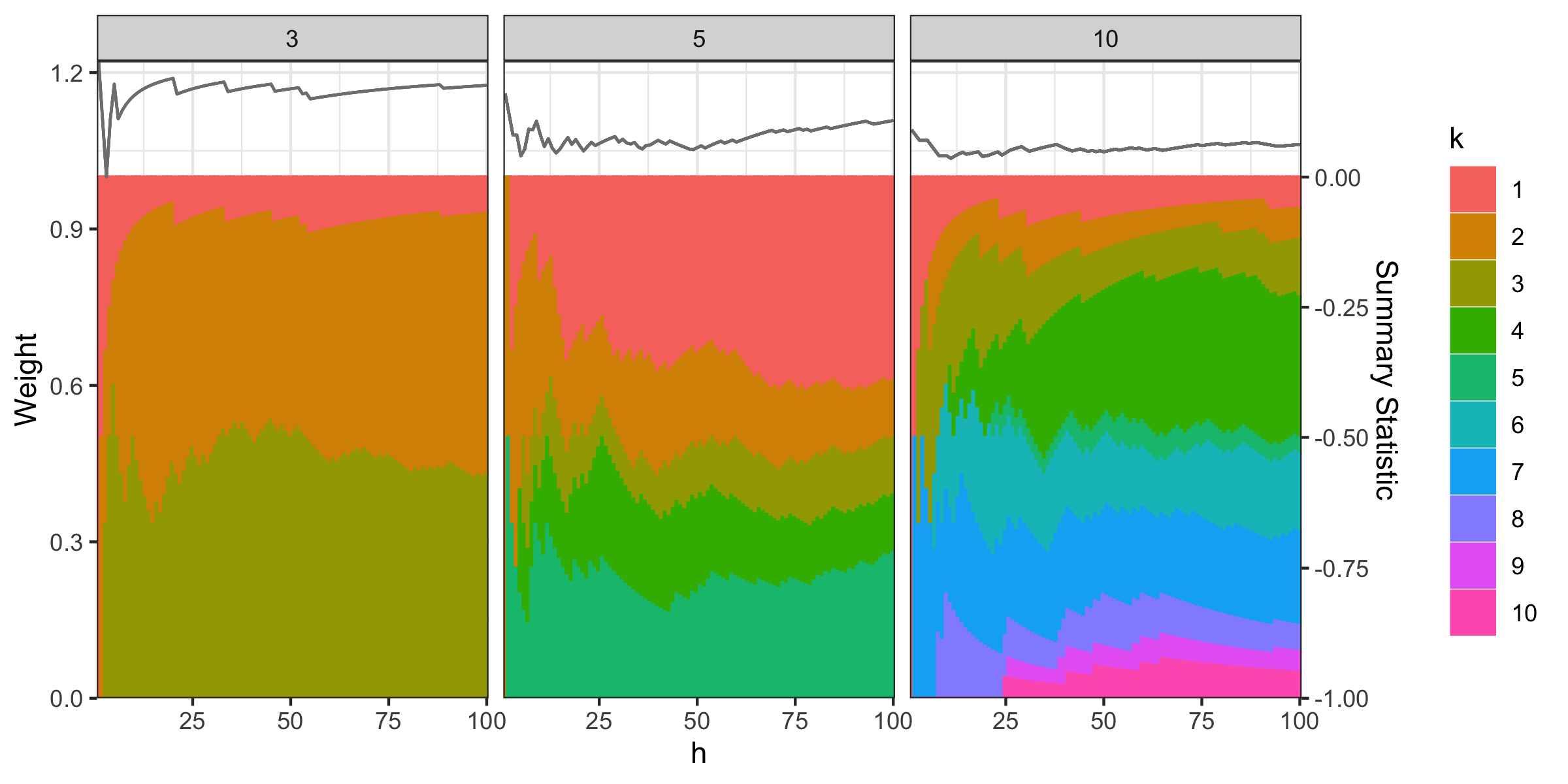}
    \caption{Hospital-based subset weights as iteration goes by in bandit selection. The experiment and visualization settings are the same as Fig. \ref{fig:all_stack_cluster0} except that the source is split randomly.}
    \label{fig:all_stack_random0}
\end{figure}

\begin{figure}[!htp]
    \centering
    \includegraphics[scale=0.06]{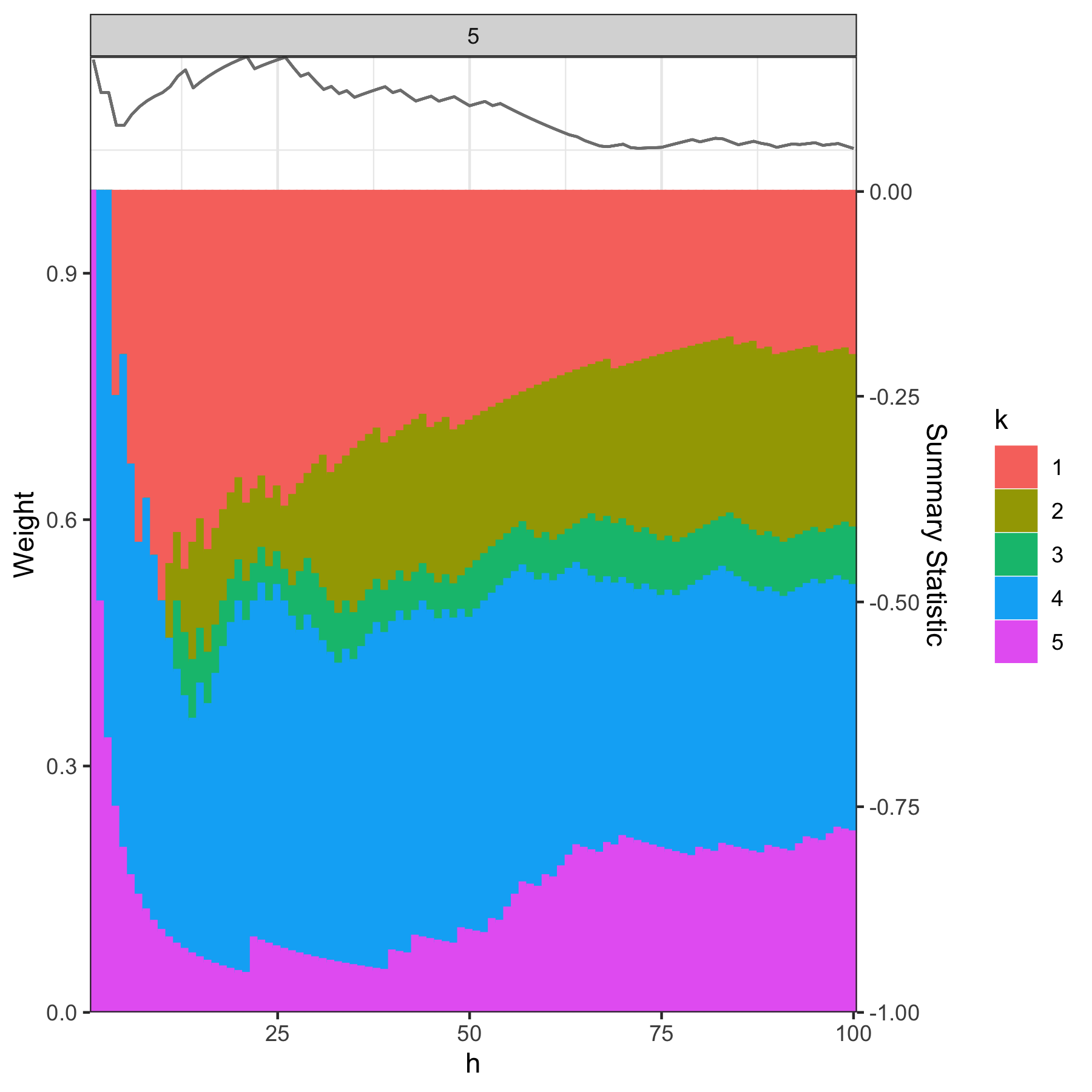}
    \caption{Hospital-based subset weights as iteration goes by in bandit selection. The experiment and visualization settings are the same as Fig. \ref{fig:all_stack_center4} except that the target hospital is hospital 1.}
    \label{fig:all_stack_center_target_0}
\end{figure}

When we use hospital 1 as target and data from all hospitals as source and perform a bandit selection, there isn't a distinguishable difference between learned mixtures and a random sample with respect to generalization performance, regardless of whether we are using deep or shallow features, or how many subsets we split the source into (Fig. \ref{fig:all_lines_target_0}). The behavior of summary statistic between these two scenarios over iterations  (Fig. \ref{fig:all_stack_cluster0} and \ref{fig:all_stack_random0}) is also similar: it goes down after a high initial value due to randomness, and then decreases after the model converges since none of the subsets are selected. We also notice that when we use hospitals to split the source, the bandit selection isn't dominantly selecting data from hospital 1 (Fig. \ref{fig:all_stack_center_target_0}). This suggests that making predictions on hospital 1 doesn't require such a source data selection under this setting. 

\begin{figure}[h]
    \centering
    \includegraphics[scale=0.09]{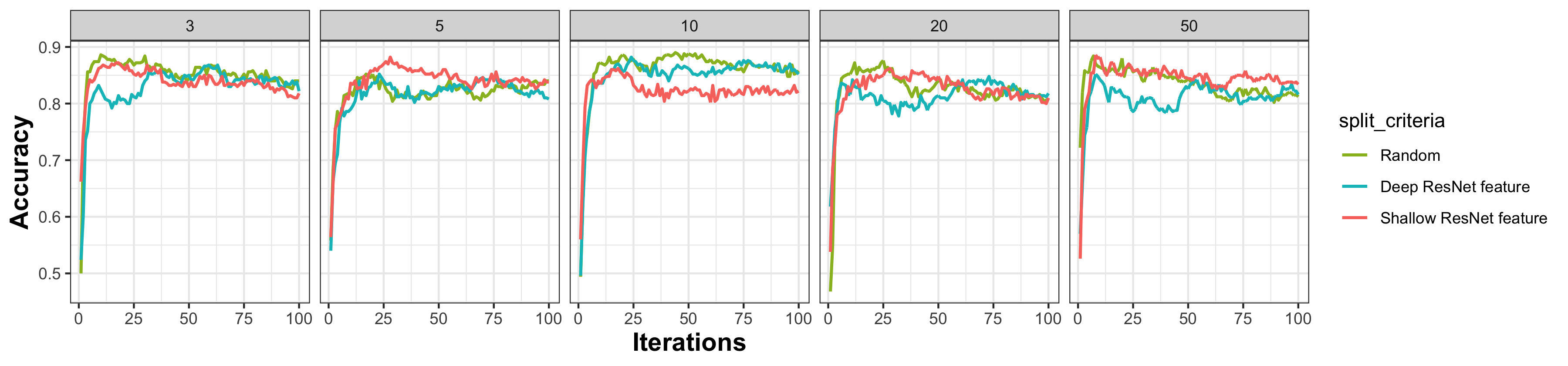}
    \caption{Model accuracy over bandit selection. The experiment and visualization settings are the same as Fig. \ref{fig:all_lines_target_0} except that we exclude data points of hospital 1 from the source.}
    \label{fig:other_lines_target_0}
\end{figure}

\begin{figure}[htp]
    \centering
    \includegraphics[scale=0.09]{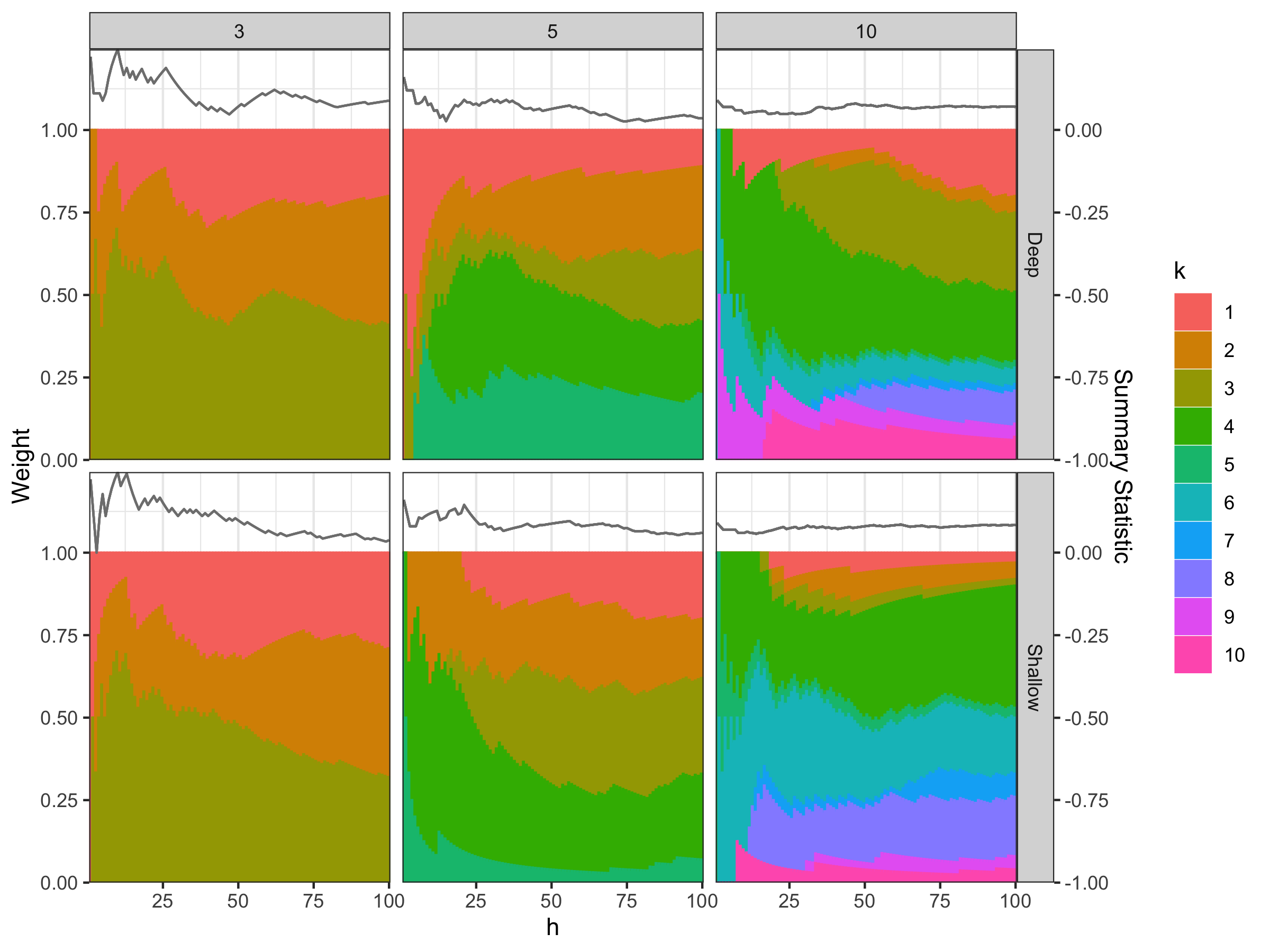}
    \caption{Hospital-based subset weights as iteration goes by in bandit selection. The experiment and visualization settings are the same as Fig. \ref{fig:all_stack_cluster0} except that we exclude data points of hospital 1 from the source.}
    \label{fig:other_stack_cluster_target_0}
\end{figure}

\begin{figure}[htp]
    \centering
    \includegraphics[scale=0.09]{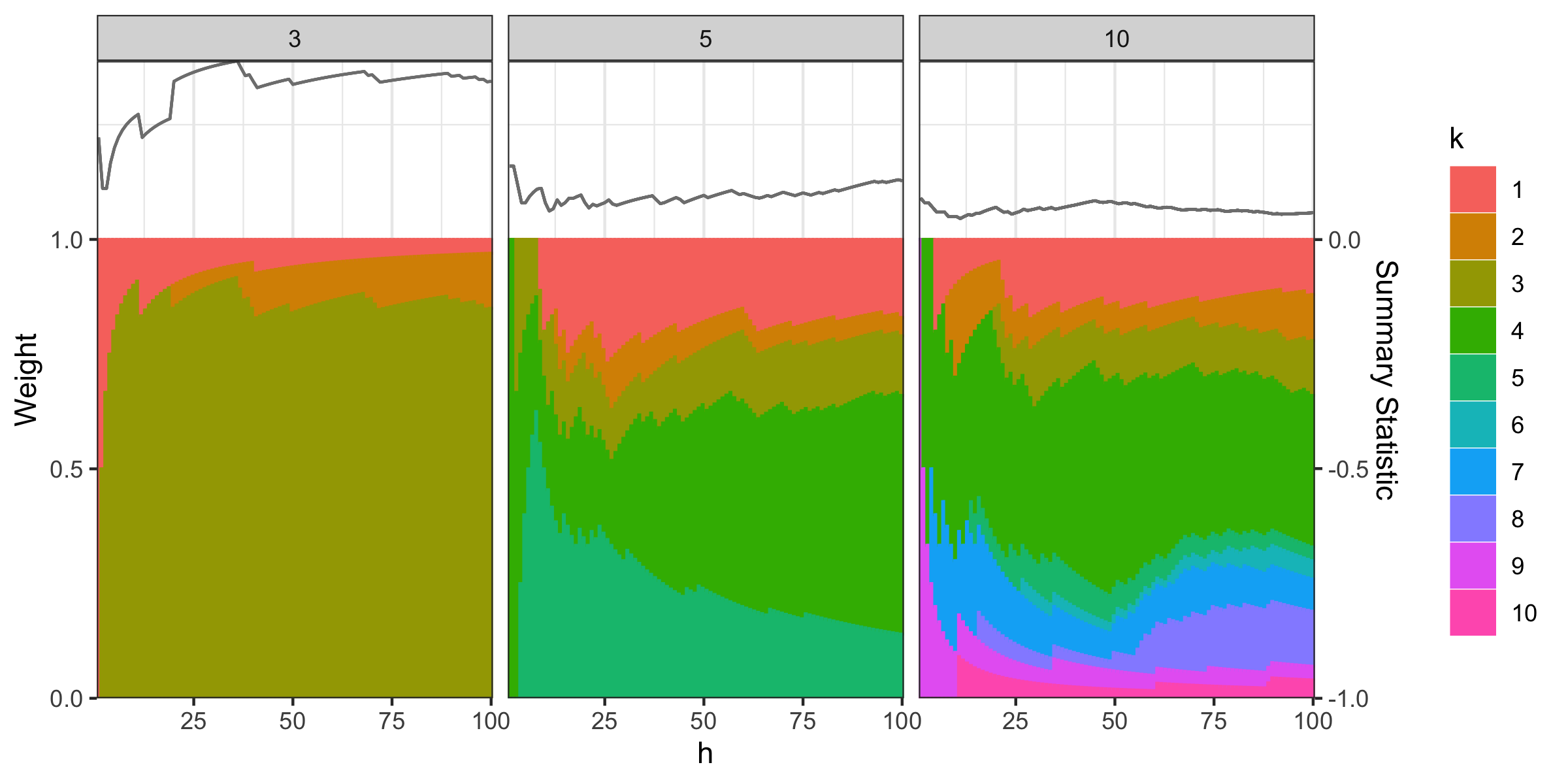}
    \caption{Hospital-based subset weights as iteration goes by in bandit selection. The experiment and visualization settings are the same as Fig. \ref{fig:other_stack_cluster_target_0} except that we split the source randomly.}
    \label{fig:other_stack_random_target_0}
\end{figure}

Furthermore, we apply a source / target split where the source excludes data from hospital 1, as what we have done for hospital 5 in section \ref{subsec:camelyon17}. There isn't an obvious performance drop (Fig. \ref{fig:other_lines_target_0}). The summary statistic and source subset weights also behave similar as the previous experiment (Fig. \ref{fig:other_stack_cluster_target_0} and Fig. \ref{fig:other_stack_random_target_0}). This confirms with the idea that in a diagnostic framework, we will conclude that source selection isn't that effective when our target is hospital 1.


\subsection{Transitional study}
\begin{figure}
    \centering
    \includegraphics[scale=0.07]{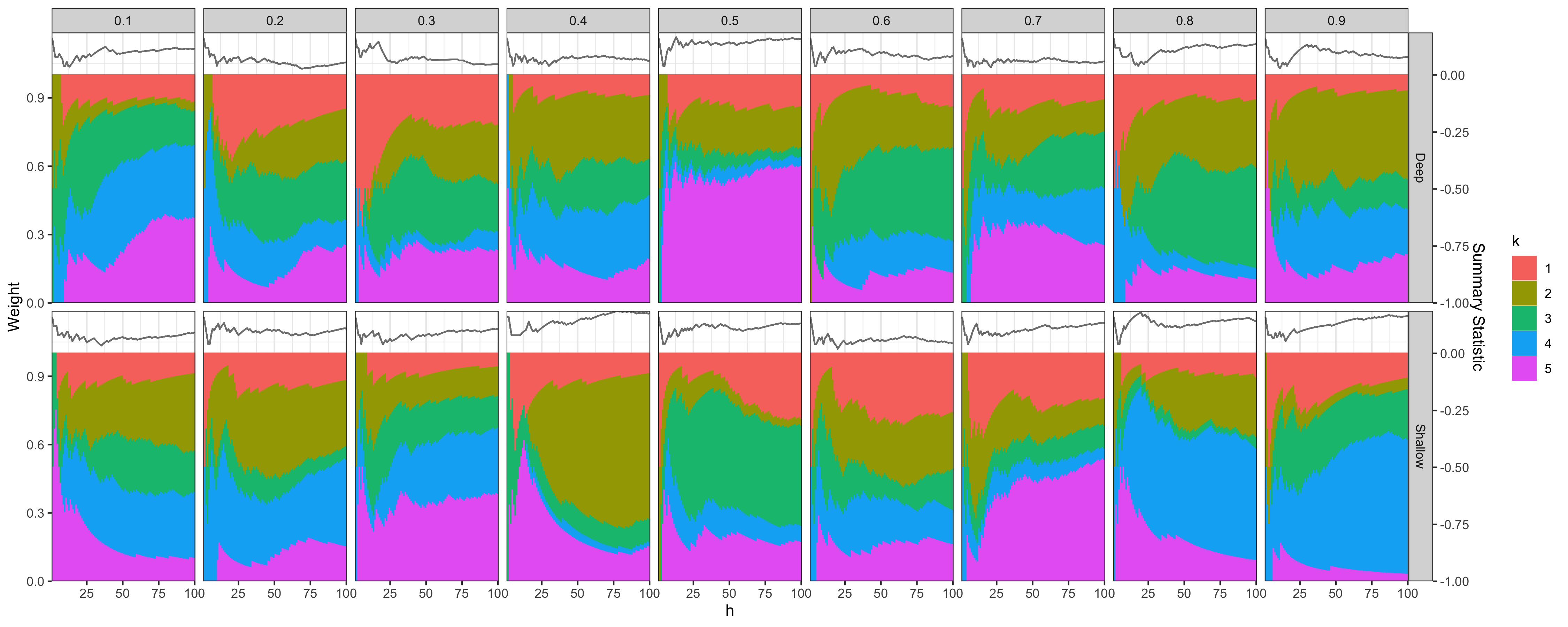}
    \caption{Transitional study for cluster-based source split. The experiment and visualization settings are the same as Fig. \ref{fig:transition_stack_center} except that the source is split by clustering (top) deep or (shallow) ResNet representations.}
    \label{fig:transition_stack_cluster}
\end{figure}

\begin{figure}[htp]
    \centering
    \includegraphics[scale=0.07]{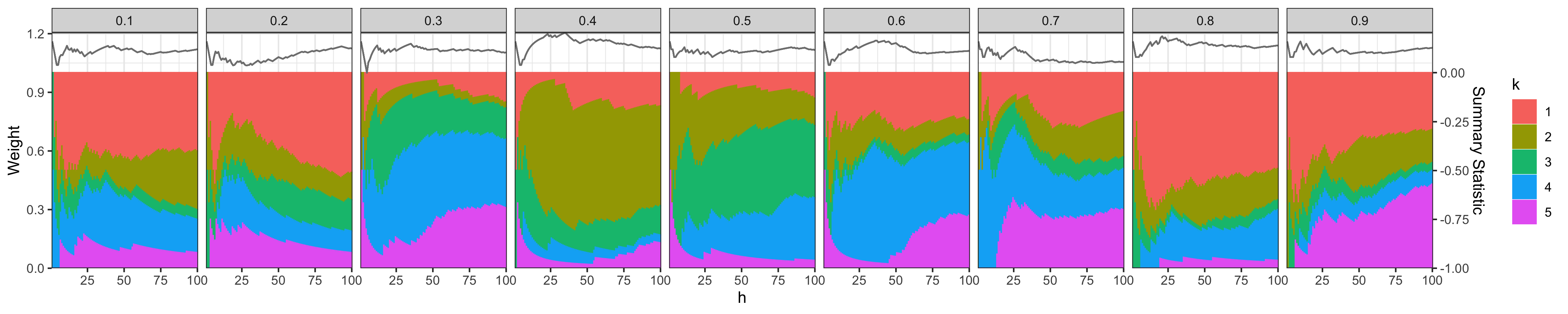}
    \caption{Transitional study for randomly-split source. The experiment and visualization settings are the same as Fig. \ref{fig:transition_stack_center} except that the source is split randomly.}
    \label{fig:transition_stack_random}
\end{figure}

As supplementary figures for the transitional study of Camelyon17-wilds dataset in section \ref{subsubsec:transition}, Fig. \ref{fig:transition_stack_cluster} and Fig. \ref{fig:transition_stack_random} show the source subset weights over the bandit selection when source is split by clustering ResNet features and randomly, respectively.

\end{appendices}

\newpage

\bibliography{bibliography}  

\end{document}